\documentclass[10pt,twocolumn,letterpaper]{article}

\usepackage{wacv}              

\usepackage{graphicx}
\usepackage{amsmath}
\usepackage{amssymb}
\usepackage{booktabs}
\usepackage{comment}
\usepackage{multirow}
\usepackage{xcolor, colortbl}
\usepackage{makecell}
\usepackage[bottom]{footmisc}

\usepackage{float}
\usepackage[accsupp]{axessibility}
\usepackage[pagebackref,breaklinks,colorlinks]{hyperref}

\usepackage[capitalize]{cleveref}
\crefname{section}{Sec.}{Secs.}
\Crefname{section}{Section}{Sections}
\Crefname{table}{Table}{Tables}
\crefname{table}{Tab.}{Tabs.}

\usepackage{orcidlink}

\definecolor{mygray}{gray}{.9}

\def\wacvPaperID{708} 
\def\confName{WACV}
\def\confYear{2025}

\begin{document}

\title{Attention-based Class-Conditioned Alignment for Multi-Source \\Domain Adaptation of Object Detectors}

\author{Atif Belal$^1$, Akhil Meethal$^1$, Francisco Perdigon Romero$^2$, Marco Pedersoli$^1$, Eric Granger$^1$ \\
$^1$ LIVIA, ILLS, Dept. of Systems Engineering, ETS Montreal, Canada\\
$^2$ GAIA Montreal, Ericsson Canada\\
{\tt\small atif.belal.1@ens.etsmtl.ca, }
{\tt\small akhilpm135@gmail.com, }
{\tt\small francisco.perdigon.romero@ericsson.com, } \\
{\tt\small \{marco.pedersoli, eric.granger\}@etsmtl.ca \vspace{-.6cm}    
}
}
\maketitle

\begin{abstract}
Domain adaptation methods for object detection (OD) strive to mitigate the impact of distribution shifts by promoting feature alignment across source and target domains. Multi-source domain adaptation (MSDA) allows leveraging multiple annotated source datasets and unlabeled target data to improve the accuracy and robustness of the detection model. Most state-of-the-art MSDA methods for OD perform feature alignment in a class-agnostic manner. This is challenging since the objects have unique modality information due to variations in object appearance across domains. A recent prototype-based approach proposed a class-wise alignment, yet it suffers from error accumulation caused by noisy pseudo-labels that can negatively affect adaptation with imbalanced data.
To overcome these limitations, we propose an attention-based class-conditioned alignment method for MSDA, designed to align instances of each object category across domains. In particular, an attention module combined with an adversarial domain classifier allows learning domain-invariant and class-specific instance representations.
Experimental results on multiple benchmarking MSDA datasets indicate that our method outperforms state-of-the-art methods and exhibits robustness to class imbalance, achieved through a conceptually simple class-conditioning strategy. 
Our code is available at: \url{https://github.com/imatif17/ACIA}.
\end{abstract}

\section{Introduction}
\label{sec:introduction}
Given the recent advancements in deep learning, OD models have demonstrated impressive results on many benchmark problems ranging from closed-world fixed class settings to open-world scenarios \cite{detr, detic-zhou-2022, yolov7-wang-2023, groundingdino-liu-2023}. However, their performance typically degrades when there is a domain shift between the training (source) and test (target) data distributions. To address this challenge, several unsupervised domain adaptation (UDA) methods \cite{adaptive_teacher, strong_weak_alignment, MT_graph, MTunbiased} have been proposed to improve OD performance in the presence of a domain shift.
Conventional UDA methods assume that the source data comes from a single distribution. In real-world scenarios, however, multiple source datasets may be available for adaptation with considerable domain shifts among these datasets due to variations in capture conditions, like weather, geographic location, and illumination. This setting is referred to as Multi-Source Domain Adaptation (MSDA). Leveraging source data from multiple distributions results in enhanced adaptation, increased generalization, and improved robustness against domain shifts\cite{moment, MDAN}. However, MSDA is challenging as it requires addressing discrepancies among different sources, along with sources and target discrepancies.

Inspired by UDA methods, the common MSDA approach involves learning task-specific knowledge from the labeled source datasets, simultaneously trying to learn a domain-invariant feature representation that generalizes well to the target domain. Several MSDA methods have been proposed for classification \cite{msda-classif-neurips-2020, adversarial_msda-neurips-2018, coupled-training-msda}, yet MSDA for OD remains relatively unexplored. This is due to 
the challenges of establishing a good trade-off between learning domain-invariant features while preserving task-specific knowledge \cite{tradeoff-da} when performed both at the image-level and instance-level (for OD), compared to only at the image-level (for classification).

\begin{figure*}
\centering
\begin{subfigure}{.33\textwidth}
  \centering
  \includegraphics[width=.99\linewidth]{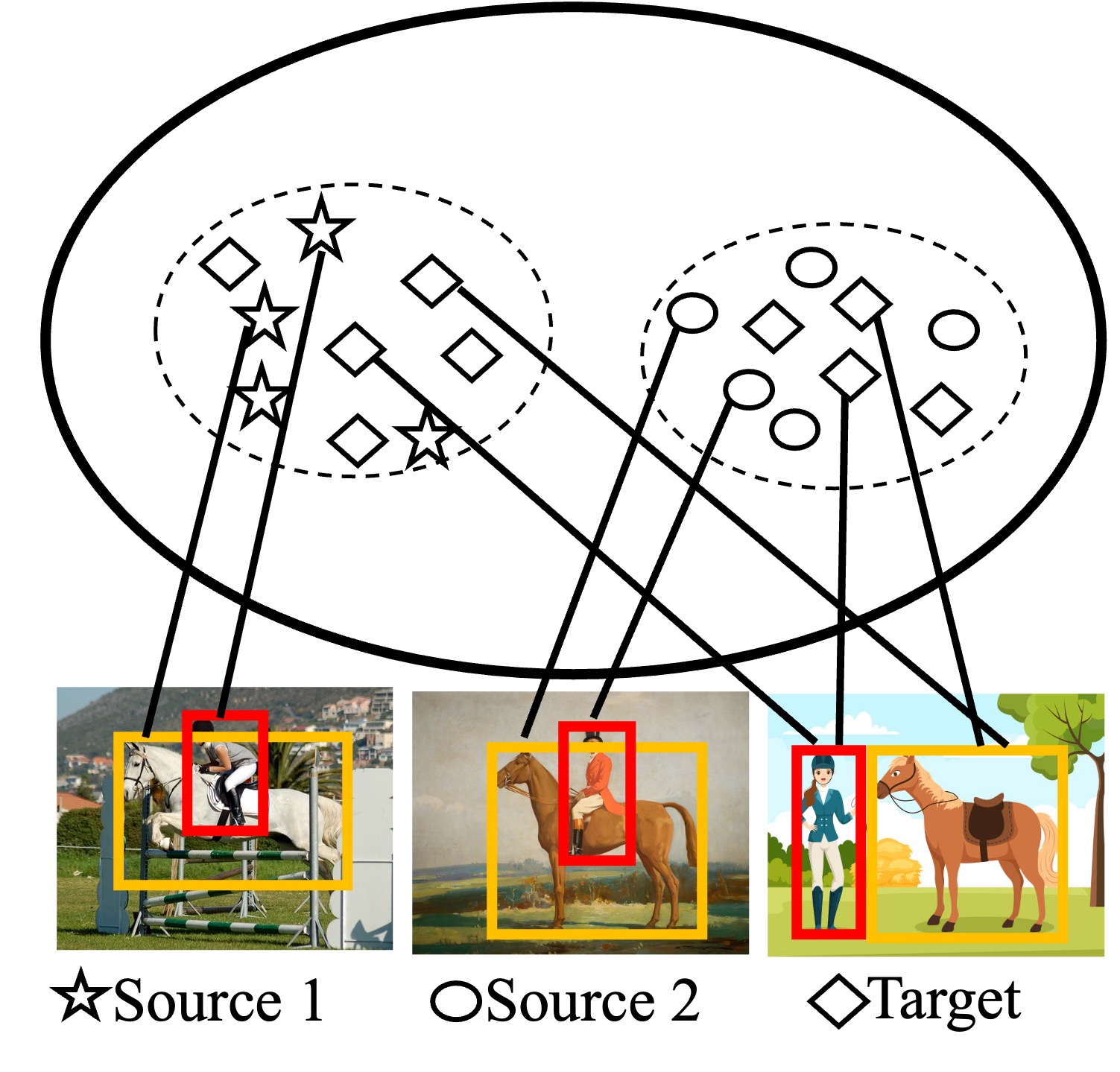}
  \caption{Methods w/o class alignment \cite{dmsn,trkp}.}
  \label{fig:sub1}
\end{subfigure}%
\begin{subfigure}{.33\textwidth}
  \centering
  \includegraphics[width=.99\linewidth]{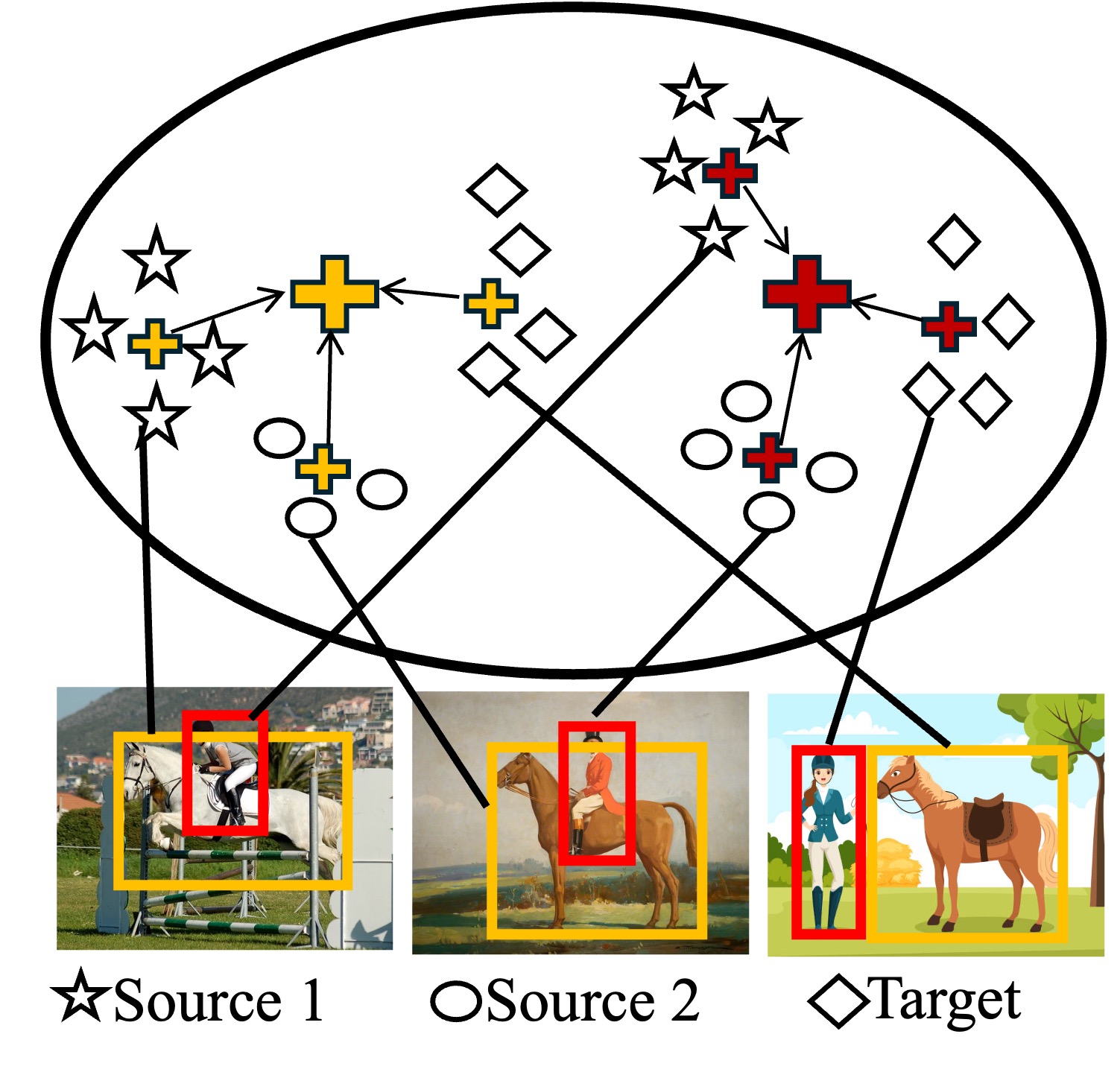}
  \caption{Prototype-based method \cite{pmt}.}
  \label{fig:sub2}
\end{subfigure}%
\begin{subfigure}{.33\textwidth}
  \centering
  \includegraphics[width=.99\linewidth]{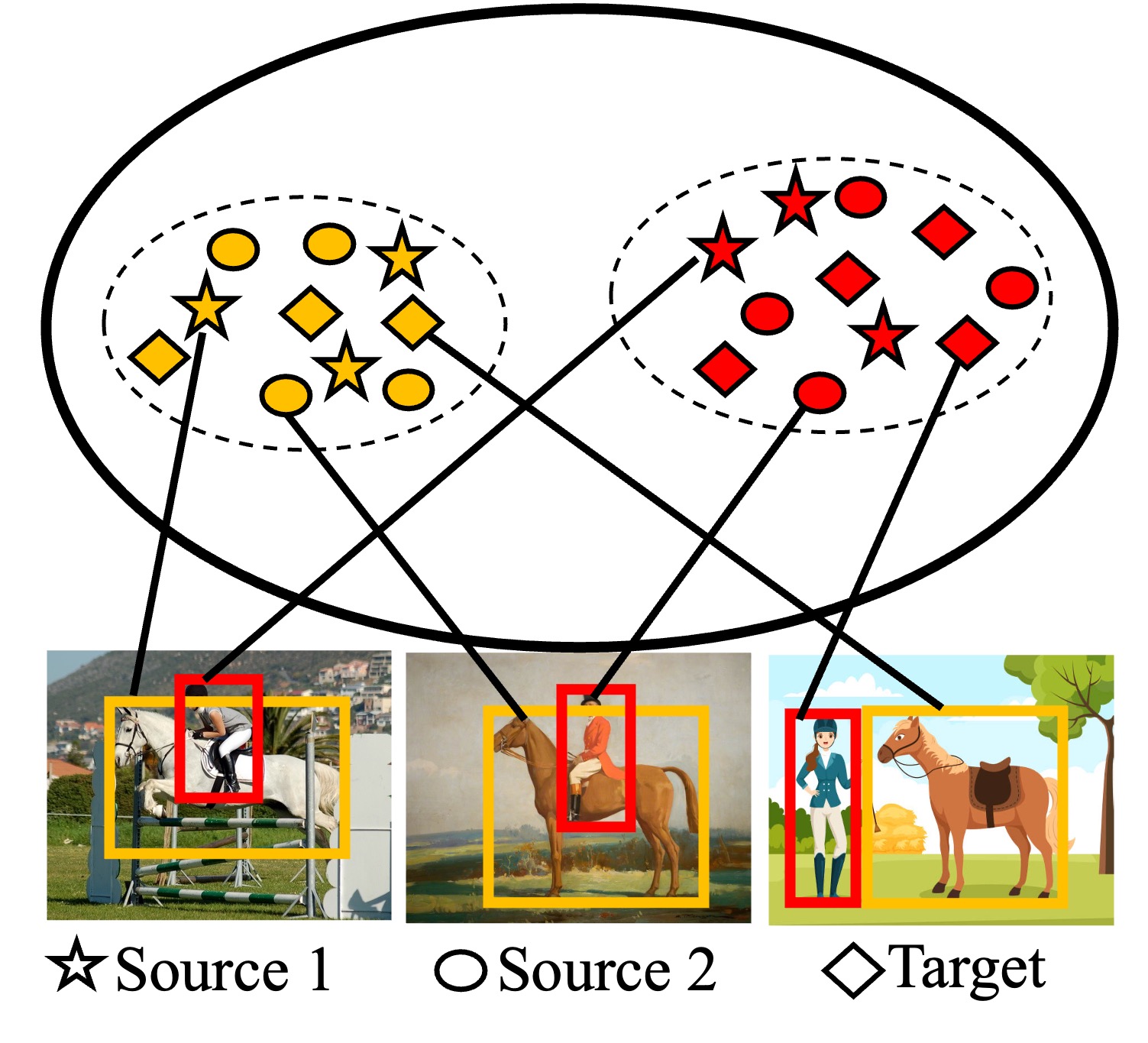}
  \caption{Our ACIA method.}
  \label{fig:sub3}
\end{subfigure}
\caption{A comparison of alignment strategies of different MSDA methods. \textbf{(a)} DMSN and TRKP implement pairwise alignment of source-target pair without considering class-wise alignment. \textbf{(b)} PMT learns specific prototypes to represent each class and domain. Then, the same class prototypes from different domains are merged into class-conditioned domain-invariant prototypes. \textbf{(c)} In contrast, our ACIA learns instance-level domain invariant features by conditioning an attention module to attend a given class.}
\label{fig:msda_align_comparison}
\vspace{-.4cm}
\end{figure*}

\begin{figure*}
\centering
\begin{subfigure}{.33\textwidth}
  \centering
  \includegraphics[width=.99\linewidth]{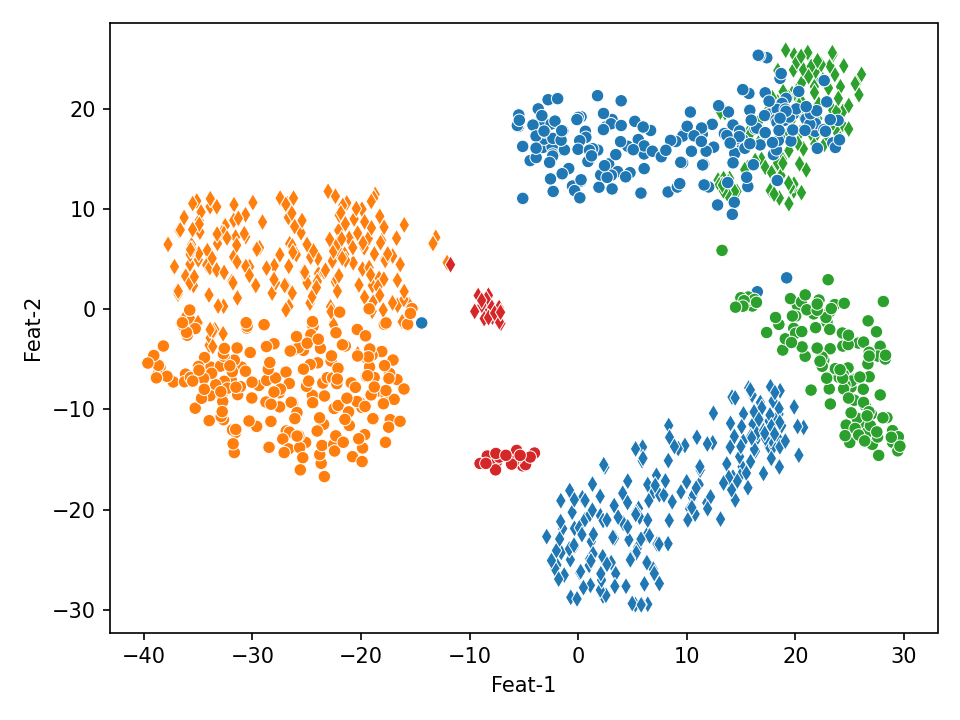}
  \caption{Methods w/o class alignment \cite{dmsn,trkp}.}
  \label{fig:tsne1}
\end{subfigure}%
\begin{subfigure}{.33\textwidth}
  \centering
  \includegraphics[width=.99\linewidth]{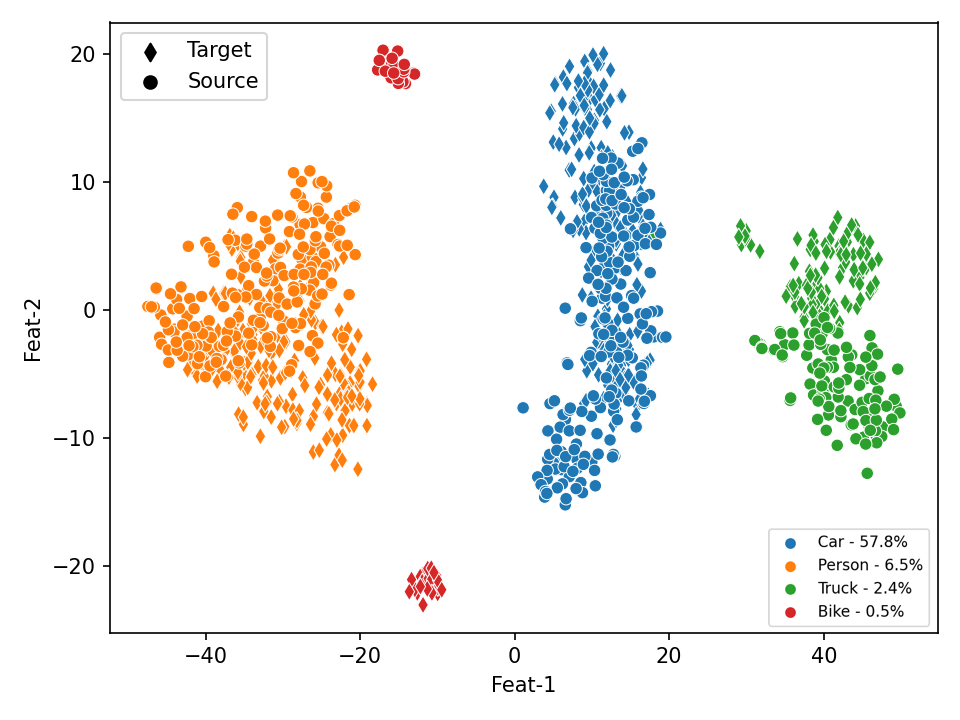}
  \caption{Prototype-based method \cite{pmt}.}
  \label{fig:tsne2}
\end{subfigure}%
\begin{subfigure}{.33\textwidth}
  \centering
  \includegraphics[width=.99\linewidth]{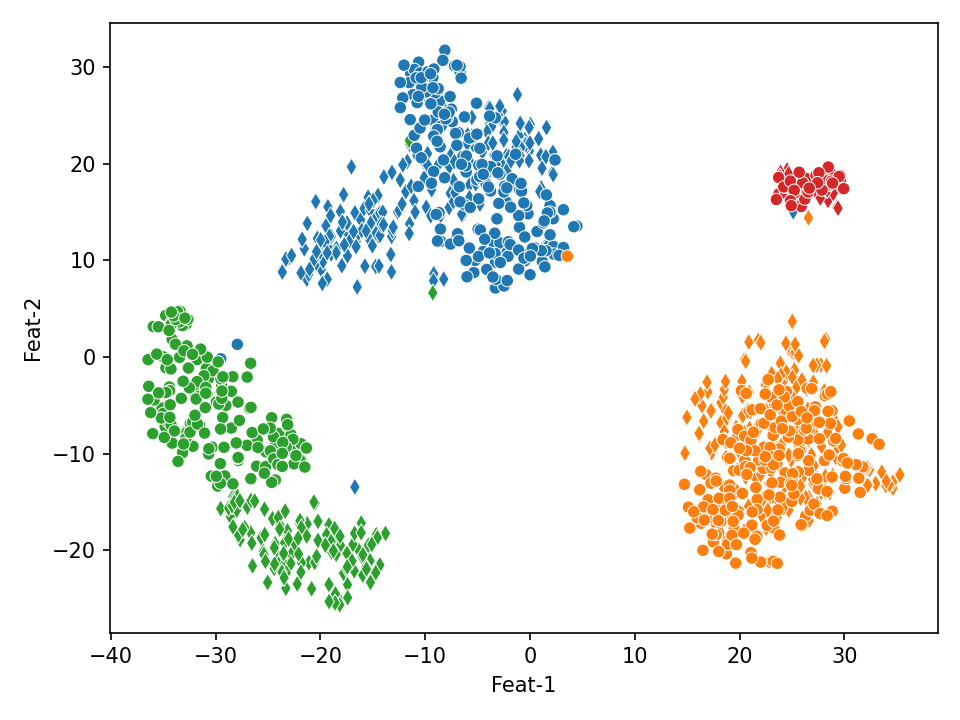}
  \caption{Our ACIA method.}
  \label{fig:tsne3}
\end{subfigure}
\caption{T-SNE projection of the class distributions of source data (BDD100K Daytime) and target data (BDD100K Dusk/Dawn). \textbf{(a)} Projection without class-specific alignment as in DMSN and TRKP. The classes are not aligned between the two domains. \textbf{(b)} PMT relies on prototypes for class alignment. Most classes are well aligned except for the bike since it is underrepresented. \textbf{(c)} Our ACIA uses an attention-based adversarial class alignment and manages to align all classes.}
\label{fig:tsne_comparison}
\vspace{-.6cm}
\end{figure*}

Only three methods have been proposed for MSDA of OD - Divide and Merge Spindle Network (DMSN) \cite{dmsn}, Target Relevant Knowledge Preservation (TRKP) \cite{trkp}, and Prototype Mean Teacher (PMT) \cite{pmt}. DMSN \cite{dmsn} and TRKP \cite{trkp} focus on pairwise alignment between each source and target pairs as shown in \cref{fig:sub1}. This ensures the alignment of each source domain with the target domain. However, in an MSDA setting, there are considerable variations in object appearances across domains, so it's also essential to focus on class alignment \cite{graph_induced, pmt} to mitigate class misalignment across domains, as can be observed in \cref{fig:tsne1}\footnote{Since both methods do not provide their code or trained model, we used our method without instance-level alignment to generate t-SNE plots.}. Later, PMT \cite{pmt} was proposed that uses prototypes for class-conditioned alignment. It relies on class-specific local and global prototypes to explicitly ensure within-domain and across-domain alignment, as summarized in \cref{fig:sub2}. However, this method has some important limitations. First, it assumes that a single prototype can represent the true distribution of a class. But, generally, a class representation is not isotropic, which can be observed in the \cref{fig:tsne2}. Second, it relies on noisy pseudo-labeled target domains to learn the prototypes, resulting in class-misalignment because of error accumulation from the pseudo-labels \cite{error_accumulation} (see \cref{table:no_target} (right)). This issue is particularly problematic for underrepresented classes, which have fewer samples and can lead to misaligned class representations (see the empirical result in \cref{table:Classwise_bdd100k}). This misalignment can be observed in \cref{fig:tsne2}, where the Bike class (red) is not aligned across the source and target domains. In this scenario, the global prototype would lie somewhere between these two distributions, which does not accurately represent the class distribution, and results in a decline in OD performance.

The above issues highlight the need for MSDA methods for OD with better class-conditioned instance alignment mechanisms. In this paper, an Attention-based Class-wise Instance Aligner (ACIA) method is proposed to efficiently align class instances across domains without relying on prototypical representation. The alignment process of ACIA is illustrated in \cref{fig:sub3}, where each object category is aligned across domains. Instances of each ROI-pooled feature are combined with the class information using an attention operation. The class information is learned by using a label embedding. To compute the attention output, the query and the value are obtained from ROI-pooled features, and the key is obtained from label embedding. The output of the attention module is fed to a domain discriminator through the Gradient Reversal Layer (GRL) \cite{GRL}. This reverses the gradient of the output of the instance-level features, forcing them to be simultaneously domain-invariant and class-wise aligned. ACIA relies on just a single class embedding to store relative class information for each object category across domains. This allows for simple and efficient multi-source class-conditional alignment. The benefits of our class-conditioned aligner can also be observed in \cref{fig:tsne_comparison}, where ACIA  provides better class alignment across domains and robustness to imbalanced data.

\noindent \textbf{Our main contributions are summarized as follows.}\\
\textbf{(1)} An MSDA method for OD called ACIA is proposed that integrates a conceptually simple class-wise instance alignment scheme. It implements instance alignment with an attention mechanism, where instances from all domains are aligned to the corresponding class.\\ 
\textbf{(2)} The combination of attention and adversarial domain classifier provides instance representations that are simultaneously domain-invariant and class-specific. \\
\textbf{(3)} An extensive set of experiments is conducted on three benchmark MSDA settings to validate the proposed ACIA alignment and characterize its key components. ACIA achieves state-of-the-art performance on all three benchmarks while remaining robust when adapting to imbalanced data. It is the most cost-effective MSDA method for OD.

\section{Related Work}  
\label{sec:related-work}

\noindent \textbf{(a) Unsupervised Domain Adaptation.} 
UDA methods aim to learn a model that performs well on the target domain by combining the information from labeled source data and unlabeled target data. A common approach for UDA is using an adversarial discriminator to generate domain-invariant representation. \cite{wildDA} integrated a discriminator based on GRL (reverses the gradient during backpropagation) at the image and instance level to align the features across domains. Later, \cite{Weak-strong} showed that strong and weak alignment of high- and low-level features can improve the UDA performance. \cite{selective} focuses on aligning the parts of the images that are more important using clustering and adversarial learning. \cite{coursetofine} argues that aligning foreground regions is more important than background regions. They used the output of RPN as an attention map and prototypes for instance-level alignment. \cite{progressive, HTCN, styletransfer} tried to mitigate domain shift at the input level. They used Cycle-GAN to convert the source to the target domain and minimize the domain shift.  
Although these works focus on a single source, they have provided important mechanisms to develop MSDA methods where source data incorporates multiple distribution shifts.

\noindent \textbf{(b) Multi-Source Domain Adaptation.} 
MSDA methods seek to adapt a model by using samples from multiple labeled source datasets and unlabeled target data. Only three works have been proposed to solve the problem of MSDA in OD -- DMSN \cite{dmsn}, TRKP \cite{trkp}, and PMT \cite{pmt}. \cite{dmsn} learns pseudo-subnets for each domain and uses adversarial alignment to align each source subnet with the target. The weight of the target subnet is obtained as the weighted exponential average of the student subnets. \cite{trkp} proposed an architecture with one detection head for each source domain. These detection heads are trained by using adversarial disentanglement for cross-domain alignment. A mining strategy is also used to weigh each source image based on its relevance to the target domain. Later \cite{pmt} propose to use class and domain-specific prototypes, which are aligned by using contrastive loss to achieve class-conditioned adaptation. Out of the three methods, only \cite{pmt} focuses on class-wise alignment. However, they rely on the assumption that class representations are isotropic, and they suffer from pseudo-label error accumulation \cite{error_accumulation}, especially when the dataset is imbalanced. In contrast, our ACIA relies on an attention-based adversarial aligner that is immune to pseudo-label error accumulation and robust to imbalanced datasets.

\noindent \textbf{(c) Instance-Level Alignment in Domain Adaptation.} 
Instance-level alignment is applied in the UDA setting to ensure domain-invariant features at the instance level. \cite{wildDA} uses an adversarial domain classifier at the instance level similar to the image-level discriminator. \cite{MCAR, categorical} improved on the work of \cite{wildDA} by giving weight to each ROI-pooled feature ensuring that more importance is given to hard-to-classify features. \cite{selective} uses a COGAN to regenerate the ROI-pooled features and later align them using adversarial training. These works align the instance-level features in a class-agnostic way. \cite{megacda} uses several memory bank prototypes to store class information across domains and align them using class-specific discriminators. 
\cite{graph_induced} learns graph-induced prototypes and proposes a loss to align them. \cite{coursetofine} used prototypes and proposed to do instance-level alignment in a class-wise fashion using a prototype-based method. Nevertheless, all these methods rely on learning prototypes for instance-level alignment in OD, based on the assumption that class representation is isotropic and uses pseudo-labels for alignment.



\section{Proposed Method}
\label{sec:proposal}
\begin{figure*}[!h]
	\centering 
		\includegraphics[width=.925\textwidth]{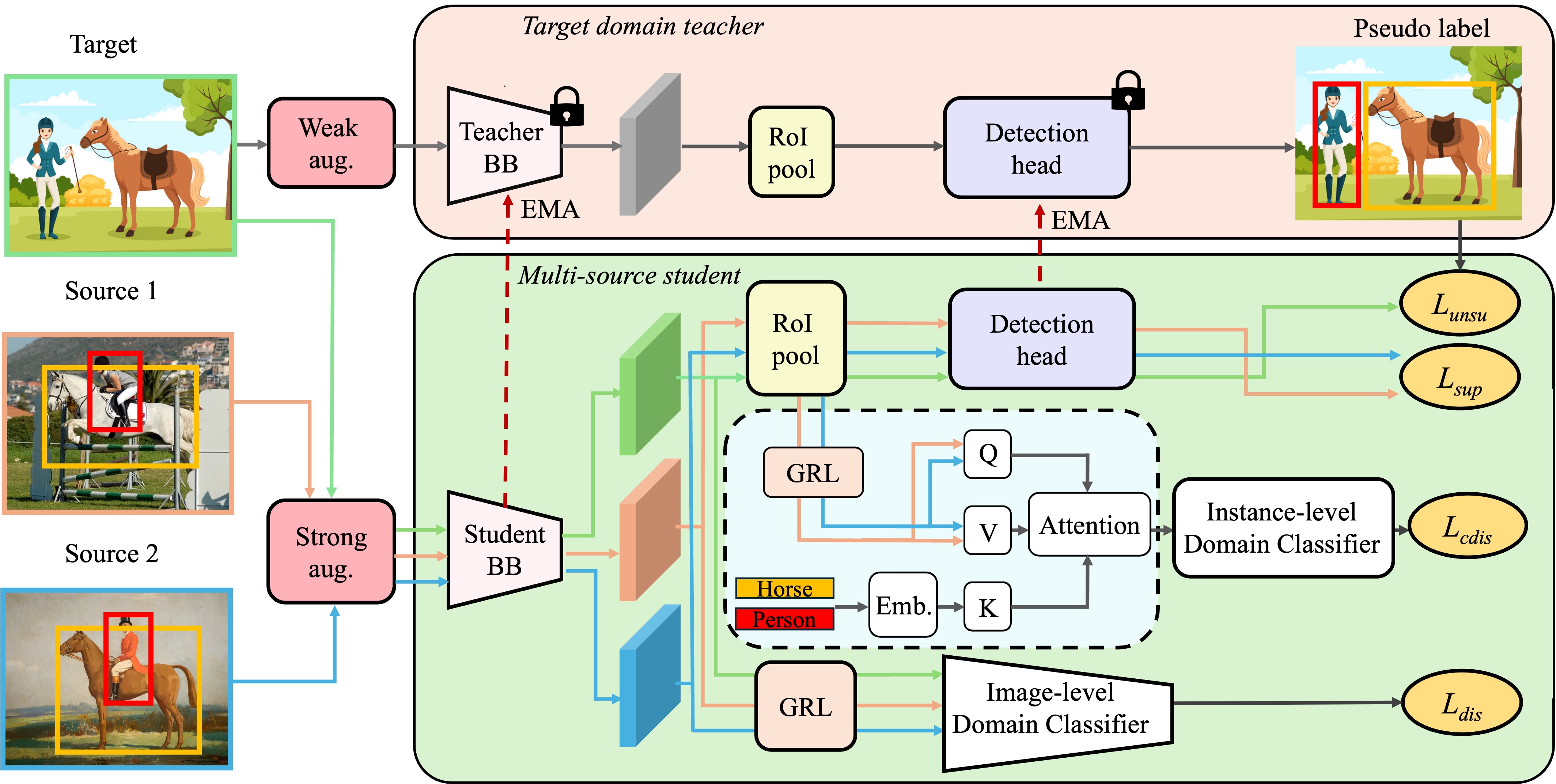}
     \\ \parbox{.9\textwidth}{\caption{
     Overview of the training architecture of our ACIA. The overall architecture is a mean-teacher model in which the student learns from multiple sources of data. An image-level classifier is introduced to globally align features, and an instance classifier to align instances in a class-conditional way, through a transformer-based attention block (see white boxes).} \label{fig:DA_training}}
    \vspace{-1.cm}
\end{figure*}

The MSDA setting consists of $N$ source domains $S_1, S_2, …, S_N$ and a target domain $T$. 
The object categories across all domains are fixed to a pre-defined set $K$. Each source domain $S_j$ consists of images and corresponding labels, denoted as $S_j = \{(x^j_i, y^j_i)\}^{M_j}_{i=1}$, where $j  \in \{1, N\}$, and $M_j$ is the number of images in the source domain $j$. Here, $x^j_i$ is the $i$th image from source domain $S_j$, and $y^j_i$ is the set of object labels and corresponding bounding boxes (bboxes). The target domain $T$ only has images with no annotations. It can be represented as $T = {\{x_i^T\}^{M_T}_{i=1}}$, where $M_T$ is the number of images in the target domain and $x_i^T$ is the $i$th image from the target domain. 

The training framework of our proposed ACIA is shown in \cref{fig:DA_training}. Our approach builds on the Mean-Teacher (MT) framework \cite{MTunbiased}, integrating components for both image-level and instance-level alignment. Specifically, these components include a multi-source discriminator for domain-invariant feature learning and an attention-based instance-level discriminator for class-wise alignment.

The MT training process is comprised of two stages. Firstly, the student model is trained in a supervised manner on all source domain data, typically referred to as the burn-in stage. Following this, the teacher model is initialized as an exact copy of the student model. Mutual learning of the teacher and student begins, utilizing both the labeled source datasets and unlabeled target data. The supervised training continues for the source datasets. Pseudo labels are generated by the teacher model to train on the target dataset. 
Meanwhile, the multi-source discriminator and the class-conditioned instance-level discriminator are trained to ensure domain-invariant features and class-wise alignment. Weights of the teacher model are the exponential moving average (EMA) of the student model weights. 
At inference time, only the teacher model is used. 

The rest of the section describes all the loss functions for the proposed ACIA framework.

\noindent \textbf{(a) Source Data Training.} 
Data from all the source domains are annotated. Thus, the standard supervised training is applied on the source datasets where the student network weights are updated through backpropagation of the detection loss: 
\begin{equation} \label{eq:super}
     \mathcal{L}_{\mbox{sup}} = \sum_{j=1}^{N}\sum_{i=1}^{M_j}\mathcal{L}_{\mbox{cls}}(x^j_i, y^j_i) + \mathcal{L}_{\mbox{reg}}(x^j_i, y^j_i) \ ,
\end{equation}
where $\mathcal{L}_{\mbox{reg}}$ is the smooth-L1 loss \cite{fasterrcnn} used for the bboxes regression and $\mathcal{L}_{\mbox{cls}}$ is the focal loss \cite{retinanet-focal-loss} used for bboxes classification. 

\noindent \textbf{(b) Target Data Training.} 
The target dataset is used at the teacher-student mutual learning stage of the training process. For every image from the target domain, two augmented versions are generated -- with strong and weak augmentation. For the strong augmentation, color jittering, grayscale, Gaussian blur, and cutout patches are used. For weak augmentation, image rescaling, and horizontal flips are used. The weak augmented image is fed to the teacher model, and the output is filtered by confidence thresholding \cite{MTunbiased} to generate pseudo-labels $\widetilde y$. 
The student network then receives the strongly augmented version and its loss is computed using the the same loss function as \cref{eq:super}:
\begin{equation}\label{eq:unsuper}
	\begin{aligned}
	\mathcal{L}_{\mbox{unsup}} =\sum_{i=1}^{M_T}\mathcal{L}_{\mbox{cls}}(x^T_i, \widetilde{y}_i) + \mathcal{L}_{\mbox{reg}}(x^T_i, \widetilde{y}_i)
	\end{aligned}
\end{equation}

\noindent \textbf{(c) Image-Level Alignment.} 
In the UDA setting, task-specific knowledge is learned from the source domain. However, for the model to perform well on the target domain, a domain-invariant representation must also be learned. This is achieved by attaching a domain discriminator (binary source-target classifier) to the output of the backbone that uses a GRL \cite{adaptive_teacher}. This results in a \textit{min-max} optimization game between the backbone and the domain-discriminator, leading to domain-invariant feature representation. In the MSDA setting, the source data comes from multiple distributions, and therefore, a multi-class discriminator is required. In our work, we used a CNN-based classifier that outputs a domain prediction map from the last feature map of the backbone. The cross-entropy loss is then applied to all pixels of the prediction map, and the results are pooled using global average pooling. The training loss of the image-level discriminator can be expressed as: 
\begin{equation}\label{eq:disc}
	\begin{aligned}
	\mathcal{L}_{\mbox{dis}} = - \frac{1}{WH} \sum_{j = 1}^{N+1} \sum_{i = 1}^{M_j} \sum_{h=1,w=1}^{H, W} d_i^j  \log(D_M(F(x_i^j)_{wh}) 
	\end{aligned}
\end{equation}
where W and H are the width and height of the prediction map. $D_M$ and $F$ represent the multi-class discriminator and feature extractor (backbone network), respectively. $D_M(F(x_i^j)_{wh}$ is the output of the multi-class discriminator for pixel $w, h$ of the prediction map. $d_i^j$ is a one-hot vector with ${N+1}$ dimensions, where the position corresponding to the domain of the $i$th image is set to 1, and 0 elsewhere.

\noindent \textbf{(d) Instance-level Alignment.} 
The instance-level alignment of ACIA is performed in a class-wise manner. Each object category has variations in visual attributes (such as appearance, scale, orientation, and other characteristics) across domains. If the variation is significant, it is challenging for the detector to align features across domains, leading to sub-optimal performance \cite{graph_induced, pmt}. Furthermore, class-wise alignment helps with underrepresented classes by ensuring that the features of these classes are explicitly aligned across domains, regardless of their lower frequency in the data. As shown in \cref{fig:tsne_comparison}, without class-wise alignment, the separation of class features of the same category across domains is high. PMT \cite{pmt} is the first MSDA approach in OD that implements class-wise alignment using global and local prototypes. The local prototypes ensured within-domain alignment, whereas the global prototypes ensured across-domain alignment. In contrast to using prototypes for each object category and each domain, we use an efficient attention-based module to ensure class-wise instance alignment.

The instance-level alignment in OD is performed using the RoI-pooled features from each instance. Unlike image-level alignment, each feature at the instance level represents only a single object category. Following previous approaches, an adversarial discriminator on RoI-Pooled features can perform the instance-level feature alignment. But this alignment process is class-agnostic. If we can perform this adversarial learning class-wise, the resulting feature alignment will be class-conditioned. For the source data, class information is available from their annotations. For the target domain, we can use the filtered pseudo-labels generated by the teacher model. However, the pseudo-labels are noisy, even after filtering. Using them would lead to an accumulation of errors during training from the noise in pseudo-labels \cite{error_accumulation}. In addition, we empirically found that using the pseudo-labels in class-conditioned alignment harms model performance (see \cref{table:no_target}).

To use class information during the alignment process, we create learnable embeddings for each object category. Each embedding is responsible for learning class information corresponding to an individual object category. The class information from embeddings can be coupled with region embeddings from the RoI-Pool (resulting in class-aware region features) using simple operations like concatenation or multiplication. It is observed that the attention mechanisms are effective at finding relations between two vectors by dynamically focusing on different parts of the sequences \cite{atten1, atten2}. In our ACIA, we employ an attention-based alignment operation. (The results from the empirical comparison are available in \cref{table:no_target}.) 
For this, two linear projections of the ROI-Pooled features and a linear projection from the label embedding are created. The projections from ROI-Pooled features are used as query $Q$ and value $V$, while the projection from label embedding is used as key $K$ for the attention block (see \cref{fig:DA_training}). The attention block uses multi-head attention as proposed in the original paper \cite{attention_is_all} which can be expressed as:
\begin{equation}\label{eq:attention}
	\begin{aligned}
	\mbox{Attention} ( Q,K,V ) = \mbox{softmax} \left(\frac{QK^T}{\sqrt{l_K}} \right) V \, 
	\end{aligned}
\end{equation}
where $l_K$ is the size of the key $K$. The dot-product of $K$ and $Q$ allows our model to learn the relation between the object features and class labels based on object category and image domain. Later, the dot-product of this with $V$, allows our model to learn which part of the object feature to focus on for class-conditioned adaptation. The multi-head attention allows our method to learn multiple such relations. At the end of the attention block, a classification head is used to predict the domain of the input instance. The instance aligner is trained with the cross-entropy loss:
\begin{equation}\label{eq:cdis}
	\begin{aligned}
	\mathcal{L}_{\mbox{cdis}} = - \sum_{j = 1}^{N} \sum_{i = 1}^{M_i} \frac{1}{Z_i^j} \sum_{r=1}^{Z_i^j} d_i^j  \log(D_I(R(F(z_{i, r}^j)))) \ ,
	\end{aligned}
\end{equation}
where $D_I$, $R$, and $F$ represent the instance-level discriminator, ROI-pooling, and feature extractor of the student model, $Z_i^j$ is the number of Ground Truth (GT) instances in the $i$th image from domain $j$, and $z_{i, r}^j$ is the $r$th GT box on the $i$th image from domain $j$. $d_i^j$ is a one-hot vector for each GT box as in \cref{eq:disc}. Note that $d_i^j$ has length $N$ because we do not consider (pseudo) GTs from the target images.

\noindent \textbf{(e) Overall Training Objective.} 
In the student-teacher mutual learning stage, all the losses discussed above are combined, and the overall training loss for the student model is:
\begin{equation}\label{eq:total_loss}
	\begin{aligned}
	\mathcal{L}= \mathcal{L}_{\mbox{sup}} + \alpha \mathcal{L}_{\mbox{unsup}} + \beta \mathcal{L}_{\mbox{dis}} + \gamma \mathcal{L}_{\mbox{cdis}} \ ,
	\end{aligned}
\end{equation}
where $\alpha$, $\beta$, and $\gamma$ are hyper-parameters for weighting losses. The teacher is updated as an EMA of the student at each iteration as $w_t = \delta w_t + (1-\delta) w_s$ where $w_t, w_s$ are the weights of teacher and student models respectively, and $\delta$ is a smoothing factor.
\section{Experiments}
\label{sec:experiments}

\subsection{Experimental Methodology} \label{sec:methodology}

\noindent \textbf{(a) Implementation Details.} 
In our experiments, we follow the same settings as \cite{dmsn, trkp, pmt}. In all experiments, Faster-RCNN \cite{fasterrcnn} is used as the detector, with Imagenet pre-trained VGG-16 backbone. Images are resized such that the shorter side of the image is 600 while maintaining the image ratios, following the ROI-alignment \cite{maskrcnn} based implementation of Faster-RCNN. We set $\alpha$, $\beta$, and $\gamma$ hyperparameters to 1, 0.1, and 0.3, respectively. The pseudo-label threshold was set to 0.7 for all the experiments. As mentioned in \cref{sec:proposal}, our training is two-staged. We trained our model for 15 epochs in both the settings. We set the weight smoothing coefficient for the EMA of the teacher model to 0.9996. All experiments were executed on 4 NVIDIA A100 GPUs, with a batch size of 8 and a learning rate of 0.2. 

\noindent \textbf{(b) Baseline Methods.} 
We compared our results for 4 settings: (1) Lower Bound: supervised only training on the blended source data, (2) UDA: source domains are blended to apply state-of-art UDA methods, (3) MSDA: state-of-art MSDA methods, and (4) Upper Bound. We consider 3 upper bounds: Target Only: supervised training on target data, All-Combined: supervised training on target and source combined, Fine-Tuning: trained on source domain, then fine-tuned on the target domain.

\noindent \textbf{(c) Datasets and Adaptation Setting.} For our experiments we use 5 different datasets with various objects and different domains and capturing conditions: BDD100K \cite{bdd100k}, Cityscapes \cite{cityscapes}, KITTI \cite{kitty}, MS COCO \cite{mscoco} and Synscapes \cite{synscapes}.
The MSDA problem settings used in the empirical study are summarized in supplementary materials. 

\subsection{Results and Discussion}
\label{sec:cross_time}

\noindent \textbf{(a) Cross-time Adaptation.}  
Daytime and Night subsets of the BDD100K dataset are used as the source domain, while the Dusk/Dawn subset of BDD100K is used as the target domain. In this setting, the images are captured at different times. This causes a domain shift due to variations in illumination.  
The performance of ACIA is compared against other baselines in \cref{table:BDD100k_sota}. Globally, UDA methods outperform the Lower Bound as expected, but there is a slight decrease in performance when the night domain is added. This is because the UDA Blending methods do not consider the domain shift among the source domains, resulting in performance degradation. This emphasizes the importance of considering MSDA methods. ACIA outperforms PMT \cite{pmt} by 2.6 mAP. Also, our method outperforms Target-Only and All-Combined Upper Bound settings. The poor performance of the Target Only setting is due to fewer samples in the target domain compared to source domains. In the all combined baseline the model is more biased towards the source datasets. Whereas our method explicitly focuses on improving the performance on the target domain that's why our result is better.
\begin{table}[!t]
    \centering
    \resizebox{.99\linewidth}{!}{
      \begin{tabular}{ c | l || c | c | c }
      \hline 
    \textbf{Setting} & \textbf{Method} & \multicolumn{3}{c}{\textbf{Source Data}} \\ 
    \textbf{ } & \textbf{ } & \textbf{  D  } & \textbf{ N } & \textbf{ D+N } \\ 
    \hline \hline
    Lower Bound & Source Only & 30.4 & 25.0 & 28.9\\
     \hline 
     
    \multirow{5}{*}{UDA} & Strong-Weak\cite{strongweak} & 31.4 & 26.9 & 29.9\\
    & Graph Prototype\cite{MT_graph} & 31.8 & 27.6 & 30.6\\
    & Cat. Regularization\cite{categorical} & 31.2 & 28.4 & 30.2\\
    & UMT\cite{MTunbiased} & 33.8 & 21.6 & 33.5 \\
    & Adaptive Teacher\cite{adaptive_teacher} & 34.9 & 27.8 & 34.6 \\
     
    \hline

    \multirow{6}{*}{MSDA} & MDAN\cite{MDAN} & - & - & 27.6 \\
    & M$^3$SDA\cite{moment} & - & - & 26.5 \\
    & DMSN\cite{dmsn} & - & -& 35.0 \\
    & TRKP\cite{trkp} & - & - & 39.8 \\
    & PMT\cite{pmt} & - & - & 45.3 \\
    \rowcolor{mygray} & \textbf{ACIA (Ours)} & - & - & \textbf{47.9} \\
    \hline

    \multirow{3}{*}{Upper Bound} &  Target-Only & - & - & 26.6 \\
    &   All-Combined & - & - & 45.6 \\
    &  Fine-Tuning & - & - & 50.9 \\
    \hline
    \end{tabular}
    }
    \vspace{-.2cm}
    \caption{mAP performance of ACIA and baseline methods for cross-time adaptation on BDD100k, when the 2 sources are daytime (D) and night (N) subsets, and the target is the \textit{Dusk/Dawn}.}
    \label{table:BDD100k_sota}
    \vspace{-.4cm}
\end{table}

\noindent \textbf{(b) Cross-camera Adaptation.}  
Cityscapes and Kitti datasets are used as source domains, while the Daytime subset of BDD100K is used as the target domain. In this setting, images are captured using different cameras, which causes a domain shift due to variations in capture conditions, like resolution and viewpoint. Training and evaluation are done only on the car category, as the only overlapping class across domains. 
The AP on the car compared against the baselines is reported in \cref{table:kitty}. The general variation of performance in this setting is similar to the Cross-time setting. Our proposed method outperforms the previous method by 0.4\%. Our result is also very close to the target-only upper-bound. In this setting there is only one object category, thus the class conditioning has no impact on instance alignment. 

\begin{table}[!t]
    \centering
    \resizebox{.99\linewidth}{!}{%
    \begin{tabular}{c | l || c | c | c }
    \hline
    \textbf{Setting} & \textbf{Method} & \multicolumn{3}{c}{\textbf{Source Data}} \\ 
    \textbf{ } & \textbf{ } & \textbf{  C  } & \textbf{ K } & \textbf{ C+K } \\ 
    \hline \hline
     Lower Bound & Source Only & 44.6 & 28.6 & 43.2\\
    
     \hline
    \multirow{4}{*}{UDA} & Strong-Weak\cite{strongweak} & 45.5 & 29.6 & 41.9\\
    & Cat. Regularization\cite{categorical} & 46.5 & 30.8 & 43.6\\
    & UMT\cite{MTunbiased} & 47.5 & 35.4 & 47.0 \\
    & Adaptive Teacher\cite{adaptive_teacher} & 49.8 & 40.1 & 48.4 \\
    \hline
    \multirow{6}{*}{MSDA} & MDAN\cite{MDAN} & - & - & 43.2 \\
    & M$^3$SDA\cite{moment} & - & - & 44.1 \\
    & DMSN\cite{dmsn} & - & -& 49.2 \\
    & TRKP\cite{trkp} & - & - & 58.4 \\
    & PMT\cite{pmt} & - & - & 58.7 \\
    \rowcolor{mygray} & \textbf{ACIA (Ours)} & - & - & \textbf{59.1} \\
    \hline
    \multirow{3}{*}{Upper Bound} & Target-Only   & - & - & 60.2 \\
        & All-Combined  & - & - & 69.7 \\
        & Fine-Tuning  & - & - & 72.1 \\
    \hline
    \end{tabular} %
    } 
    \caption{AP for the car class of ACIA and baselines methods for the cross-camera setting. Source domains are Cityscapes (C) and Kitty (K) and the target is \textit{Daytime} subset of BDD100K.}
    \label{table:kitty}
    \vspace{-.4cm}
\end{table}

\label{sec:mixed_da}
\noindent \textbf{(c) Extension to Mixed Domain Adaptation.}   
The MS COCO, Cityscapes, and Synscapes datasets are used as source domains, while the Daytime subset of BDD100K is used as the target domain. 
In this setting, domain shift occurs due to multiple reasons. 
Results in \cref{table:coco_sota} indicate that the Lower Bound outperforms UDA Blending baselines due to a large domain shift among domains. Thus, blending all the source domains as one and using UDA results in performance degradation. This highlights the difficulty of this setting. When Cityscapes and MS COCO are used as source domains, our method outperforms the previous state-of-the-art (PMT) by 2.3 mAP. When Synscapes is also added to the source domain this number rises to 2.6 mAP, indicating that our method improves even further when more source domains are added. 
\begin{table}[!t]
    \centering
     \resizebox{.49\textwidth}{!}{
      \begin{tabular}{ c | l || c| c| c }
      \hline
    \textbf{Setting} & \textbf{Method} & \multicolumn{3}{c}{\textbf{Source Data}} \\ 
    \textbf{ } & \textbf{ } & \textbf{  C  } & \textbf{ C+M } & \textbf{ C+M+S } \\ 
    \hline \hline
    Lower Bound & Source Only & 23.4 & 29.7 & 30.9\\
     \hline  
   \multirow{2}{*}{UDA} & UMT\cite{MTunbiased} & - & 18.5 & 25.1\\
    & Adaptive Teach.\cite{adaptive_teacher} & - & 22.9 & 29.6\\
    \hline
    \multirow{3}{*}{MSDA} & TRKP\cite{trkp} & - & 35.3 & 37.1 \\
    & PMT\cite{pmt} & - & 38.7 & 39.7 \\
   \rowcolor{mygray} & \textbf{ACIA (Ours)} & - & \textbf{41.0} & \textbf{42.3} \\
     \hline
   \multirow{3}{*}{Upper Bound} & Target-Only & - & - & 38.6 \\
   & All-Combined & - & 47.1 & 48.2 \\
   & Fine-Tuning & - & 49.2 & 52.5 \\
   \hline
    \end{tabular}
    }
    \vspace{-.2cm}
    \caption{mAP performance on 7 object categories of ACIA and baseline methods in the mixed setting. Source domains are Cityscapes (C), MS COCO (M), and Synscapes (S) datasets while the \textit{Daytime} domain of BDD100K is the target domain.}
    \label{table:coco_sota}
    \vspace{-.6cm}
\end{table}

\subsection{Ablation Studies}

Unless mentioned, all ablation studies are performed on the cross-time adaptation setting as in \cite{dmsn, pmt}. The impact of key components performing image-level and instance-level feature alignment is analyzed along with other possible design choices for class-conditioned instance alignment. 

\noindent \textbf{(a) Impact of the adaptation components.} 
The impact of the individual components of our proposed method is analyzed -- image-level discriminator and instance-level discriminator. To understand the importance of class conditioning in instance alignment, we examined the instance classifier with and without class-wise attention. Results are reported in \cref{table:components}. By adding the image-level discriminator, the performance improved considerably. Furthermore, by adding the instance-level discriminator (without class-embedding), the performance of the model increased by 3.4\%, showing the importance of instance-level alignment. The performance is further increased by integrating the class information into the instance-level alignment.
\begin{table}[!t]
    \centering
    \resizebox{.48\textwidth}{!}{
      \begin{tabular}{c |c| c || c }
    \hline
    \multirow{2}{*}{\textbf{Image-LA}} &   \textbf{Instance-LA} & \textbf{Instance-LA} & \multirow{2}{*}{\textbf{$\textrm{AP}_{50}$}}\\
    \textbf{} &   \textbf{w/o class-embed} & \textbf{w/ class-embed} & \textbf{} \\
    \hline \hline
     &  &  & 31.7 \\
    $\checkmark$    &               &               & 41.9 \\
    $\checkmark$    & $\checkmark$  &               & 45.3 \\
    $\checkmark$    &               & $\checkmark$  & 47.9\\
    \hline
    \end{tabular}
    }
    \vspace{-0.2cm}
    \caption{Impact of individual ACIA components. Image-LA: Image-Level Alignment, Instance-LA: Instance-Level Alignment.}
    \label{table:components}
    \vspace{-.5cm}
\end{table}

\noindent \textbf{(b) Level of imbalance.}  
In \cref{table:Imbalance_bdd100k} compares ACIA against PMT for different levels of imbalance. When the dataset is balanced, both methods perform similarly. However, when the level of imbalance increases, ACIA obtains higher mAP. This shows that the error accumulation from the pseudo-labels is more problematic with imbalanced datasets. Our class-conditioned attention model mitigates this issue because it uses the same model for all classes, and therefore can better deal with high levels of imbalance.
\begin{table*}
    \centering
    \resizebox{.85\linewidth}{!}{
    \setlength{\tabcolsep}{5pt}
      \begin{tabular}{ l || c | c | c | c | c | c | c | c | c  || c }
      \hline
    \textbf{Method }  & \textbf{Bike } & \textbf{Bus } & \textbf{Car } & \textbf{Motor } & \textbf{Person } & \textbf{Rider } & \textbf{Light } & \textbf{Sign } & \textbf{Truck } & \textbf{mAP }\\ 
    \hline
    \hline
    \textbf{\textit{Mostly Balanced}} & \textbf{25.0\%} & \textbf{50.0\%} & & \textbf{10.0\%} & & \textbf{15.0\%} & & & & \\ \hline
    PMT &  43.8  & 52.3  & -  &  28.0  &  - & 34.4  &  -   &  -  &  - & \textbf{39.6} \\
    ACIA (Ours) & 45.8  & 48.9  & -  &  30.7  &  - & 31.0  &  -   &  -  &  - & 39.1 \\
    \hline \hline
    \textbf{\textit{Slightly Imbalanced}} & \textbf{4.6\%} & \textbf{9.2\%} & & \textbf{1.8\%} & \textbf{59.6\%} & \textbf{2.8\%} & & & \textbf{22.0\%} &\\ \hline
    PMT & 42.1  & 52.7  & -  &  25.1  &  43.1 & 34.0  &  -   & -  &  50.5 & 41.2 \\
    ACIA (Ours) & 44.3  & 54.2  & -  &  25.3  &  45.3 & 40.6  &  -   & -  &  50.4 & \textbf{43.4} \\
    \hline \hline
    \textbf{\textit{Mostly Imbalanced}} & \textbf{1.2\%} & \textbf{2.4\%} & & \textbf{0.5\%} & \textbf{15.4\%} & \textbf{0.7\%} & \textbf{32.8\%} & \textbf{42.3\%} & \textbf{5.7\%} &\\ \hline
    PMT & 49.2  & 56.2  & -  &  29.6  &  49.9 & 32.0  &  43.2   & 55.8  & 55.3 & 46.4 \\
    ACIA (Ours) & 52.1  & 59.2  & -  &  29.2  &  53.1 & 36.1  &  46.2   & 58.1  & 57.1 & \textbf{48.9} \\ \hline
    \hline 
    \textbf{\textit{Highly Imbalanced}} & \textbf{0.5\%} & \textbf{1.0\%} & \textbf{57.8\%} & \textbf{0.2\%} & \textbf{6.5\%} & \textbf{0.3\%} & \textbf{13.4\%} & \textbf{17.8\%} & \textbf{2.4\%} &\\ \hline
    PMT & 55.3  & 59.8  & 67.6  &  29.9  &  47.6 & 32.7  &  46.3   & 56.0  &  57.7 & 50.3 \\ 
    ACIA (Ours) & 56.1  & 61.0  & 69.2  &  31.9  &  51.8 & 39.8  &  49.2   & 59.0  & 61.0 & \textbf{53.2} \\
    \hline
    
    \end{tabular}
    }
    \caption{Detection results per class (AP) on the cross-time setting for PMT and our method with different levels of imbalance. The dataset is quite imbalanced. To create a more balanced dataset we reduced the number of classes by removing those that are creating the imbalance. For each level of imbalance, we report the relative importance of each class in percentage.}
    \label{table:Imbalance_bdd100k}
    \vspace{-.4cm}
\end{table*}

\noindent \textbf{(c) Design choices for class-wise instance alignment,}  
The class information can be incorporated into ROI-Pooled features in different ways, like concatenation and multiplication. \cref{table:no_target} (left) reports the variation in performance with various design choices. Attention is the better design choice. We believe this is because just one class-embedding layer is learned across every domain. When merging the class information to the ROI-Pool features, attention can learn what information is more important for a particular domain in a better way than multiplication and concatenation. This is because the design of the attention layer allows it to learn complex relations \cite{attention_is_all}.
\begin{table}
\parbox{.0\linewidth}{
\centering
\resizebox{.19\textwidth}{!}{
      \begin{tabular}{l ||c}
      \hline
    \textbf{Merging} & \textbf{D+N} \\ 
    \hline \hline

    Concatenation & 46.3 \\
    Multiplication  & 46.8 \\
    Attention  & 47.9 \\
    \hline
    \end{tabular}
    }
    
}
\hfill
\parbox{.6\linewidth}{
\centering
\resizebox{.28\textwidth}{!}{
      \begin{tabular}{ l || c | c }
      \hline
     \multirow{2}{*}{\textbf{Setting}}& \textbf{With} & \textbf{Without} \\
      & \textbf{Target} & \textbf{Target} \\
    \hline \hline
     Cross-Time & 43.3 & 47.9
     \\
     Mixed & 37.9 & 41.0
     \\
     \hline
    \end{tabular}
    }
}
\caption{\textbf{(left)} Different design choices for integrating class information in instance alignment; \textbf{(right)} Detection performance degrades when the target domain is used in instance alignment.}
\label{table:no_target}
\vspace{-.8cm}
\end{table}

\noindent \textbf{(d) Impact of target domain on class-wise alignment.}  
In this section, we investigate the impact of employing or omitting the target image in instance-level alignment. As reported in \cref{table:no_target} (right), utilizing target data in both scenarios leads to a decline in detector performance. In a DA problem, the target data lack labels, making it challenging to distinguish between foreground regions and object categories. One approach to address this challenge is by employing pseudo-labels, however, they often introduce noise and degrade performance. Nonetheless, in MSDA, the availability of multiple source domains enables us to choose which domains to align at the instance level. Hence, our work focuses on aligning solely the source domains.

\begin{figure*}[]
\centering
\begin{tabular}{c @{\hspace{-0.3cm}} c  @{\hspace{-0.3cm}} c}
\includegraphics[width=0.33\linewidth, height=3.0cm]{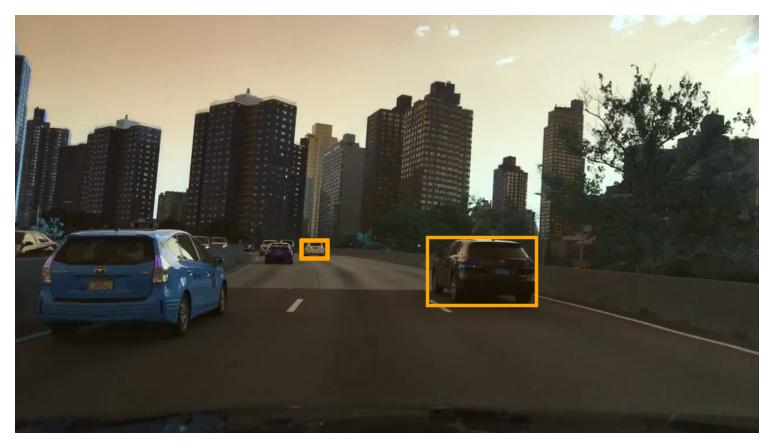} &
\includegraphics[width=0.33\linewidth, height=3.0cm]{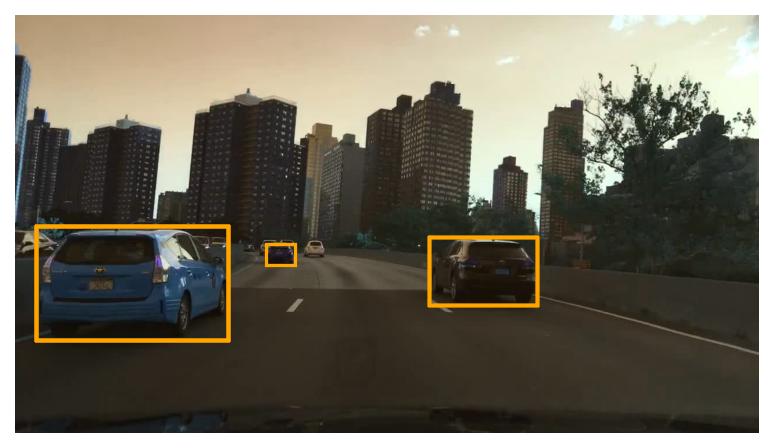} &
\includegraphics[width=0.33\linewidth, height=3.0cm]{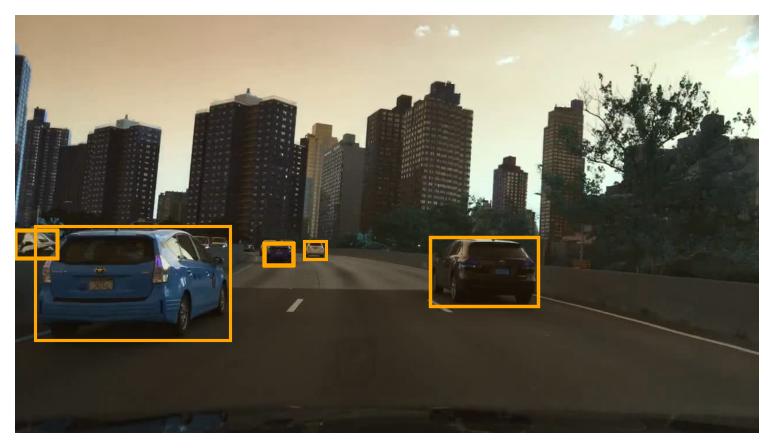} \\
\includegraphics[width=0.33\linewidth, height=3.0cm]{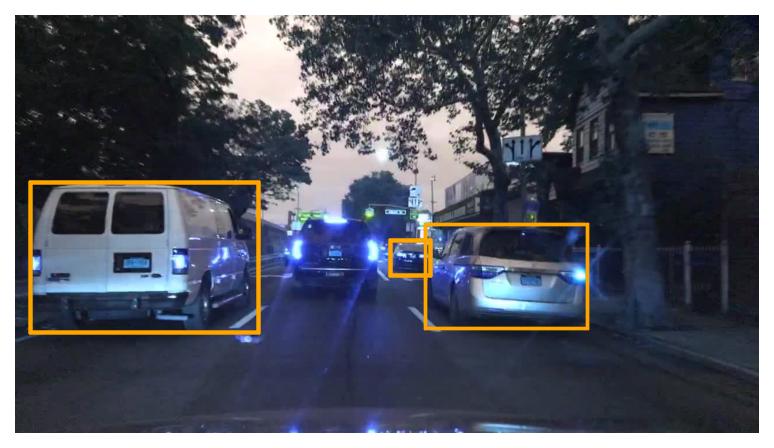} &
\includegraphics[width=0.33\linewidth, height=3.0cm]{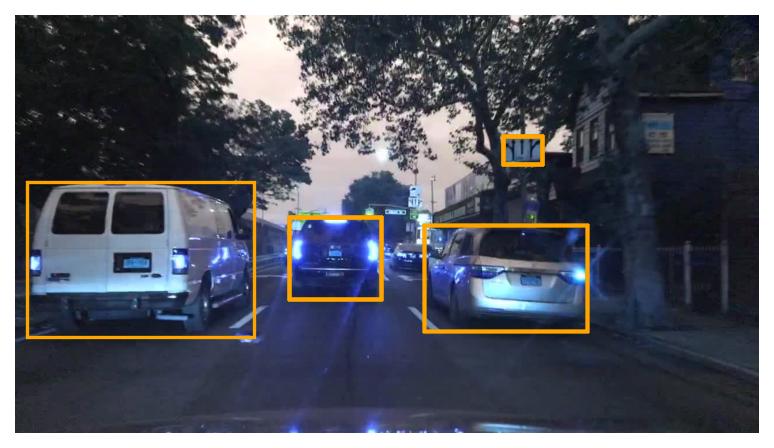} &
\includegraphics[width=0.33\linewidth, height=3.0cm]{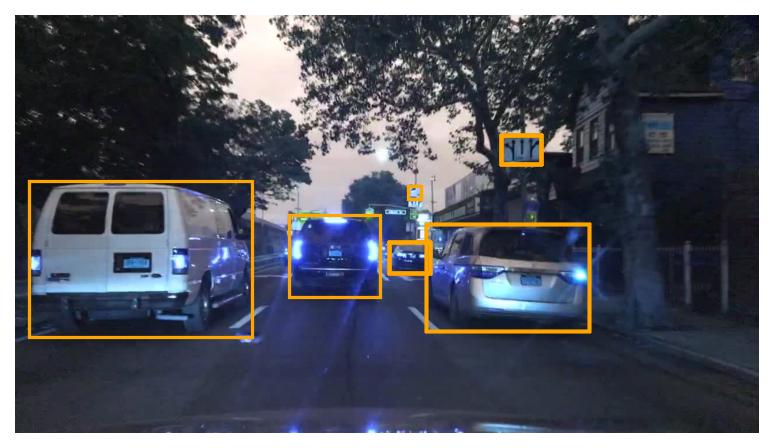} \\
\includegraphics[width=0.33\linewidth, height=3.0cm]{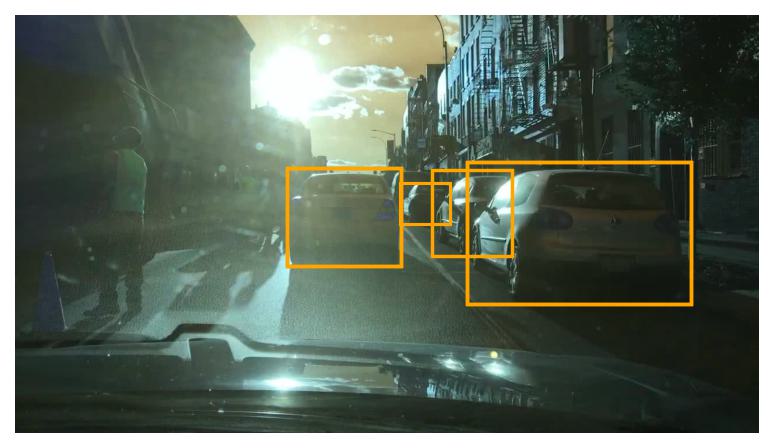} &
\includegraphics[width=0.33\linewidth, height=3.0cm]{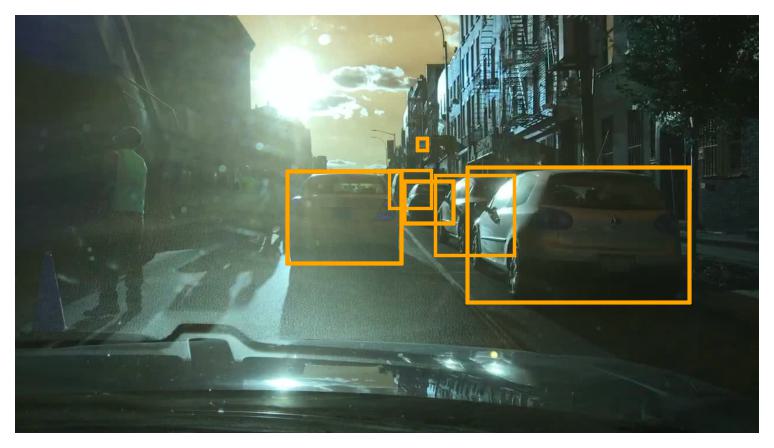} &
\includegraphics[width=0.33\linewidth, height=3.0cm]{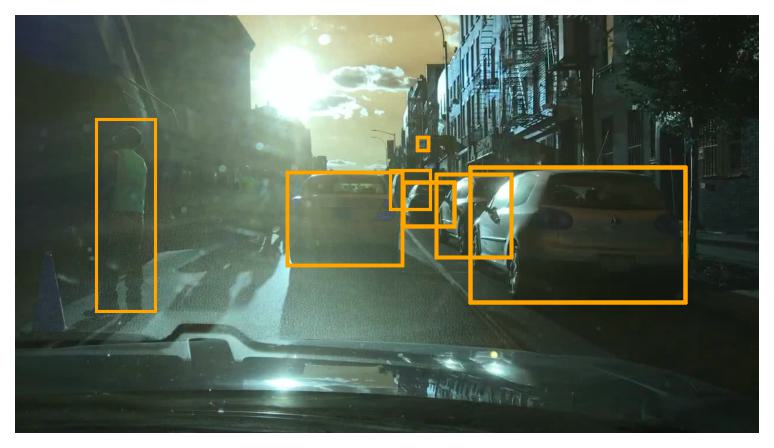} \\
 (a) Methods w/o class alignment.  & (b) Prototype-based method. & (c) Our ACIA method. \\
 \end{tabular}
 \vspace{-0.2cm}
 \caption{Examples of ODs on the BDD100k cross-time setting, with different types of instance-level adaptations. \textbf{(a)} ODs with a multi-source adaptation without instance level class-conditional adaptation as in \cite{dmsn, trkp}. \textbf{(b)} ODs with the prototype-based class-conditional adaptation \cite{pmt}. \textbf{(c)} ACIA: ODs with our attention-based class-conditional adaptation.}
 \label{fig:detections}
 \vspace{-0.5cm}
\end{figure*}

\noindent \textbf{(e) Detection visualization.}  
In \cref{fig:detections}  we present three detection visualizations from no class-wise adaptation, PMT, and our ACIA. From the first two examples, it can be observed that our ACIA is better at detecting small objects that are harder to see. In the third example, our ACIA accurately detects a person, that has fewer samples in the dataset.

\section{Conclusion} 
\label{sec:conclusion}
Recent studies in UDA have shown that class-conditioned instance alignment can improve adaptation performance \cite{graph_induced, coursetofine}. However, an analysis of class-conditioning in MSDA for OD, and its impact on instance alignment is lacking in the literature. We observed empirically that state-of-the-art class-conditioning approaches have limitations that can lead to misalignment, particularly for underrepresented classes. In this paper, we propose ACIA, an MSDA method that integrates a cost-effective class-wise instance alignment scheme. Instance alignment is implemented with an attention mechanism where instances from all domains are aligned to the corresponding class. Our ACIA method provides state-of-the-art performance on challenging MSDA benchmarks while being robust when adapting to imbalanced data. 

\noindent \textbf{Acknowledgements: } This work was supported by the Natural Sciences and Engineering Research Council of Canada and Mitacs. We also thank the Digital Research Alliance of Canada for the use of their computing resources.

\section{Supplementary Material} 

\subsection{Datasets} 
\label{dataset}
In our paper, we used five datasets to create the multi-source domain adaptation settings. These datasets are listed below and summarized in the \cref{table:msda_datasets}:

\noindent \textbf{1. BDD100k -} The BDD100K \cite{bdd100k} is a large-scale diverse driving dataset. It contains 70,000 training and 10,000 testing images captured across various times, like Daytime, Night, and Dusk/Dawn. This variation makes it a good choice for the DA problem.

\noindent \textbf{2. Cityscapes -} Cityscapes \cite{cityscapes} is an autonomous driving research dataset with images captured from urban street scenes. It contains  2,975 training and 500 testing images.

\noindent \textbf{3. Kitty -} The KITTI \cite{kitty} dataset is a self-driving dataset that comprises a collection of images and associated sensor data captured from a moving vehicle in urban environments. It consists of 7,481 training images RGB images.

\noindent \textbf{4. MS COCO -} MS COCO \cite{mscoco} is one of the most widely used benchmark datasets in computer vision. It is a complex dataset having large scale and appearance variations for the instances. It has around 330,000 images containing 80 object categories.

\noindent \textbf{5. Synscapes -} Synscapes \cite{synscapes} is a synthetic autonomous driving dataset that provides more variability for testing our method. It consists of 25,000 training images.

    \begin{table*}[h]
    \centering
      \begin{tabular}{ l || cc | cc| cc| cc}
      \hline
     \textbf{Setting} & Src.1 & \# Img. & Src.2 & \# Img.  & Src.3 & \# Img. & Target & \# Img. \\
    \hline \hline
    Cross-Time & Day & 36,728  & Night & 27,971 & - & - & Dawn & 5,027 \\
    Cross-Camera & Cityscapes & 2,831  & Kitty & 6,684 & - & - & Day & 36,728 \\
    Mixed Domain & Cityscapes & 2,975  & COCO & 71,745 & Synscapes & 25,000 & Day & 36,728 \\

   \hline
    \end{tabular}
    \caption{Summary of the different MSDA settings used in our study. 
    }
    \label{table:msda_datasets}
\end{table*}

\subsection{Study on the Class-Embedding layer}

\begin{figure}[h]
	\centering 
		\includegraphics[width=.5\textwidth]{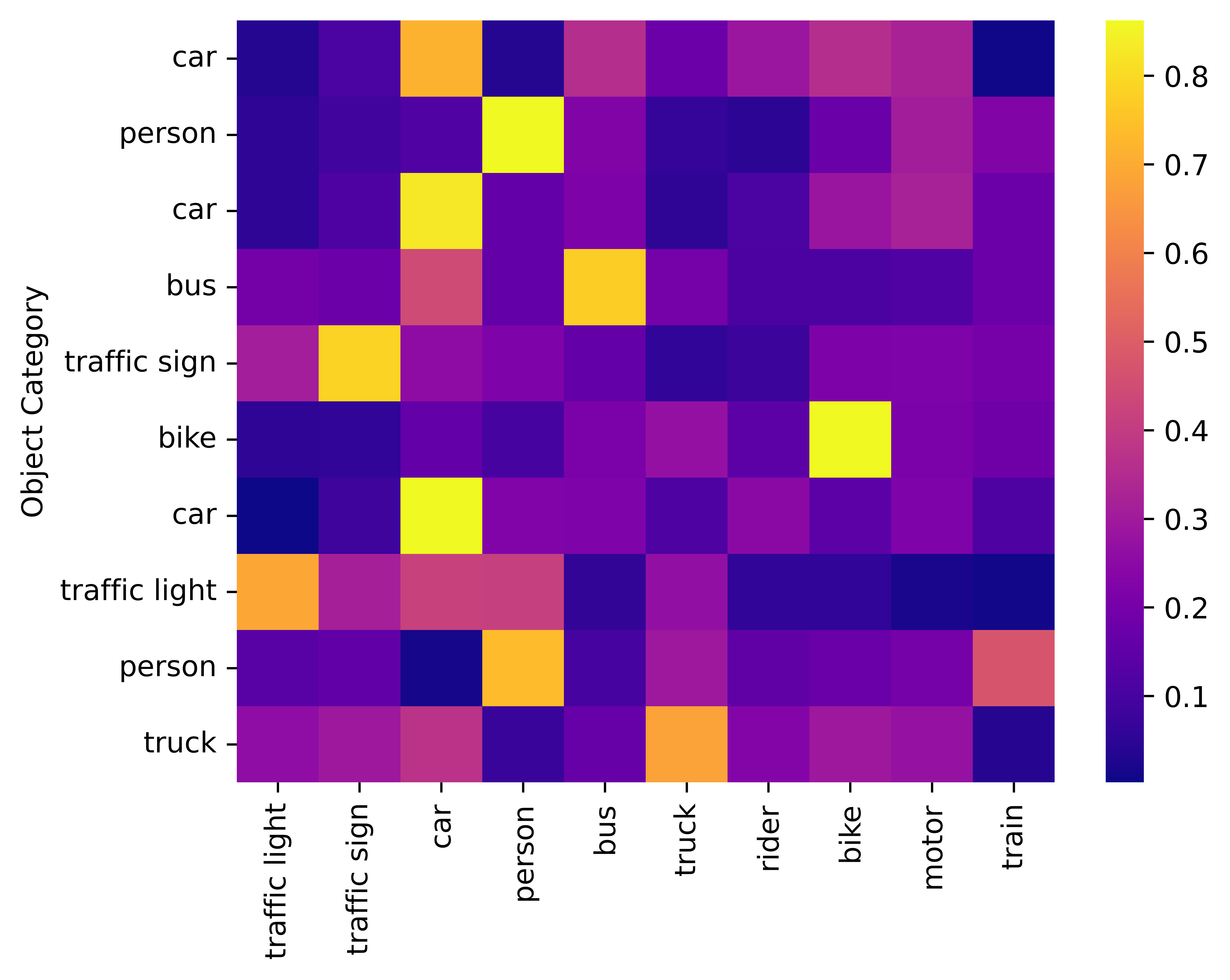}
     {\caption{
      Heatmap showing the activation of each class embedding with instance features of some object categories. The X-axis represents the list of all the class embedding corresponding to all the object categories, and the Y-axis represents the object category present in the ROI-Pooled feature.} \label{fig:heat}}
      \vspace{-.55cm}
\end{figure}

In this subsection, we visualize the information learned by the class-embedding layer. \cref{fig:heat} shows the activation of each class-embedding layer (corresponding to each object category) with ROI-Pooled features that contain the object category. For this, we crop the region of the object category from the image and find activation with each class-embedding layer. After that, we use Softmax to normalize the values. It can be observed from \cref{fig:heat} that each embedding layer is activated most when it matches the corresponding object category. This shows that the class-embedding is successfully learning class information. Also, the under-represented object categories (bike, truck, truck) are highly activated with their corresponding embedding layer. It shows that our attention-based instance-level alignment mechanism is helping in datasets with class imbalance by focusing on under-represented classes.

\subsection{Increase in number of parameters with source domains}

Earlier works in the MSDA for OD \cite{dmsn, trkp} relied on learning domain-specific parameters. It rapidly increases training parameters with each source domain, as summarized in \cref{table:param_growth}. Later, PMT \cite{pmt} mitigated this by using prototypes instead of domain-specific subnets. However, in PMT the number of parameters slightly increases with each source domain. In contrast, our ACIA doesn't have any domain-specific parameters, so there is no change in the number of parameters with source domains.

\begin{table*}[h]
    \centering
      \begin{tabular}{c || c c c c c}
      \hline
    \textbf{\multirow{2}{*}{Method}} &\multicolumn{5}{c}{Number of source domains}\\
    \cline{2-6}
    & 1 & 2 & 3 & 4 & 5\\
    \hline \hline

   DMSN & 45.994 & 75.426 & 104.858 & 134.290 & 163.722 \\
   TRKP & 45.994 & 59.942 & 73.890 & 87.838 & 101.786 \\
   PMT & 46.586 & 46.587 & 46.588 & 46.589 & 46.590\\
   \textbf{ACIA (ours)} & 45.659 & 45.659 & 45.659 & 45.659 & 45.659\\
   \hline
    \end{tabular}
    \caption{The number of parameters vs the number of source domains. No parameter growth in our method.}
    \label{table:param_growth}
\end{table*}

\subsection{Equalization Loss v2 for class-imbalance}

Focal loss was introduced as an improvement over cross-entropy loss, it helps in dealing with foreground-background imbalance. But, the MSDA setting (cross-time) used in our paper have foreground-foreground class imbalance. In this subsection, we replace the focal loss for object detection with equalization loss v2 \cite{eqlv2} to tackle the problem of class imbalance. EQLv2 loss was proposed for long-tail object detection and have demonstrated significant improvements in detecting under-represented objects without sacrificing performance on more frequent classes. Equalization Loss v2 adjusts the gradients during backpropagation to address class imbalance in object detection. It applies a class-specific weighting factor to the gradient of the classification loss. The frequent classes are down-weighted  and rare classes are up-weighted  based on their frequency, thereby balancing the learning process for all classes. In \cref{table:EQLLoss}, we compare the performance of EQL v2 loss with focal loss on the cross-time setting. We also trained PMT with this loss for comparison. It can be observed that in both the lower and upper bound, EQL v2 is outperforming focal loss. This shows the effectiveness of EQL v2 loss in a class-imbalance problem. In case of PMT, EQL v2 is improving their performance by a slight margin. This shows that PMT is not very effective in imbalance dataset scenarios.
For ACIA, there is no improvement when focal loss is replaced by EQLv2 loss. This further proves that our method is every effective when dealing with class imbalance problem.

\begin{table}[!ht]
    \centering
      \begin{tabular}{ c | l || c |c }
      \hline 
    \textbf{Setting} & \textbf{Method} & \textbf{Focal Loss} & \textbf{EQLv2}\\
    \hline \hline
    Lower Bound & Source Only & 28.9 & 31.6 \\
     \hline 
    \multirow{2}{*}{MSDA}& PMT & 45.3 & 45.7\\
     & \textbf{ACIA (Ours)} & \textbf{47.9}  & \textbf{47.5}\\
    \hline

    \multirow{3}{*}{Upper Bound} &  Target-Only  & 26.6 & 28.3\\ 
    &   All-Combined  & 45.6 & 45.8\\
    &  Fine-Tuning  & 50.9 & 51.7\\
    \hline
    \end{tabular}
    \caption{Comparison of performance when using focal loss and equalization loss v2 for OD on the cross-time setting.}
    \label{table:EQLLoss}
    \vspace{-.4cm}
\end{table}

\subsection{Additional Experiments on the Cross-Time Setting}
In this subsection, we provide some additional results on the cross-time settings. Here, we present a setting where source domain contains images which are mostly in bright environment while target domain contains dark/dull environment. For this, we used Daytime and Dusk/Dawn domain of BDD100K as source domains, while Night domain of BDD100K is used as target domain. This setting is challenging due to large domain shift between source and target domain. The result for this setting is reported in \cref{table:night_target}, we also trained PMT and reported their performance for comparison. It can be observed that, we are outperforming PMT by 2.3 mAP. This shows that our instance-level alignment performs better when the domain shift is large between source and target domain. 

\begin{table}[!ht]
    \centering
      \begin{tabular}{ c | l || c }
      \hline 
    \textbf{Setting} & \textbf{Method} & \textbf{mAP} \\
    \hline \hline
    Lower Bound & Source Only & 24.2 \\
     \hline 
    \multirow{2}{*}{MSDA}& PMT & 34.9 \\
     & \textbf{ACIA (Ours)} & \textbf{37.2} \\
    \hline

    \multirow{3}{*}{Upper Bound} &  Target-Only  & 37.8 \\
    &   All-Combined  & 42.1 \\
    &  Fine-Tuning  & 46.8 \\
    \hline
    \end{tabular}
    \caption{mAP performance of ACIA and baseline methods for cross-time adaptation on BDD100k, when the two sources are Daytime and Dusk/Dawn subsets, and the target is the \textit{Night} subset.}
    \label{table:night_target}
\end{table}

\subsection{Additional Experiments on the Mixed Domain Adaptation Setting} In this subsection, we provide some additional results on the mixed domain adaptation settings. In the main paper, we only showed the results when $C+M$ was employed as the two source domains (because the previous papers followed that setting only). In \cref{table:additional_mixed} we compare our results with PMT\footnote{We trained for this setting using the code provided by them} when $C+S$ and $M+S$ are employed as the source domain. It can be observed that our method outperforms PMT for both settings. This further shows that our attention-based instance alignment performs better than prototype-based instance alignment with complex domain shifts.

\begin{table}[H]
\centering
  \begin{tabular}{ c|| c| c}
  \hline
\textbf{Method} & \textbf{ C+S } & \textbf{ M+S } \\ 
\hline 
\hline
     PMT & 30.1 & 34.9 \\
     \textbf{ACIA (ours)} & \textbf{33.7} & \textbf{35.8} \\
\hline
    \end{tabular}
    \caption{Additional results on the Mixed Domain Adaptation Setting.}
    \label{table:additional_mixed}
\end{table}

\subsection{Effect of Class-Alignment in UDA methods}
In this subsection, we show the effect of class-wise alignment on UDA methods \cite{strong_weak_alignment, adaptive_teacher}. \cref{table:uda_class_align} shows the results. We studied three different cases: (1) No class alignment - we used their proposed method only. (2) Class-Alignment with target domain - here we incorporated our class alignment component with their method, aligning the source domains and the target domain. (3) Class-Alignment without target domain - our class alignment component with their method, but this time only the source domains are considered. The results clearly show that our class alignment is effective with both methods. Thus, we can conclude that the proposed class-conditional alignment is effective for both UDA and MSDA methods. It can be observed that similar to our method, removing target data for the instance-level alignment further enhances the model performance.
    
\begin{table}[H]
    \centering
    \resizebox{.49\textwidth}{!}{
      \setlength{\tabcolsep}{4pt}
      \begin{tabular}{c || c c c }
      \hline
    \multirow{2}{*}{\textbf{Method}} & \textbf{No Class} & \textbf{Class-Align.} & \textbf{Class-Align.}\\ 
    & \textbf{Align.} & \textbf{w/ Target} & \textbf{w/o Target}\\
    \hline
    \hline
     Strong-Weak\cite{strong_weak_alignment} & 29.9 & 33.7 & 34.2
     \\
     Adaptive Teacher\cite{adaptive_teacher} & 34.6 & 36.8 & 37.6
     \\
     \hline
    \end{tabular}
    }
    \caption{Importance of Class-alignment in UDA methods. The proposed attention-based class-conditioned aligner is effective for UDA methods as well.}
    \label{table:uda_class_align}
\end{table}

\subsection{Architecture of the Discriminator Networks}
\begin{figure}[h]
	\centering 
		\includegraphics[width=.49\textwidth]{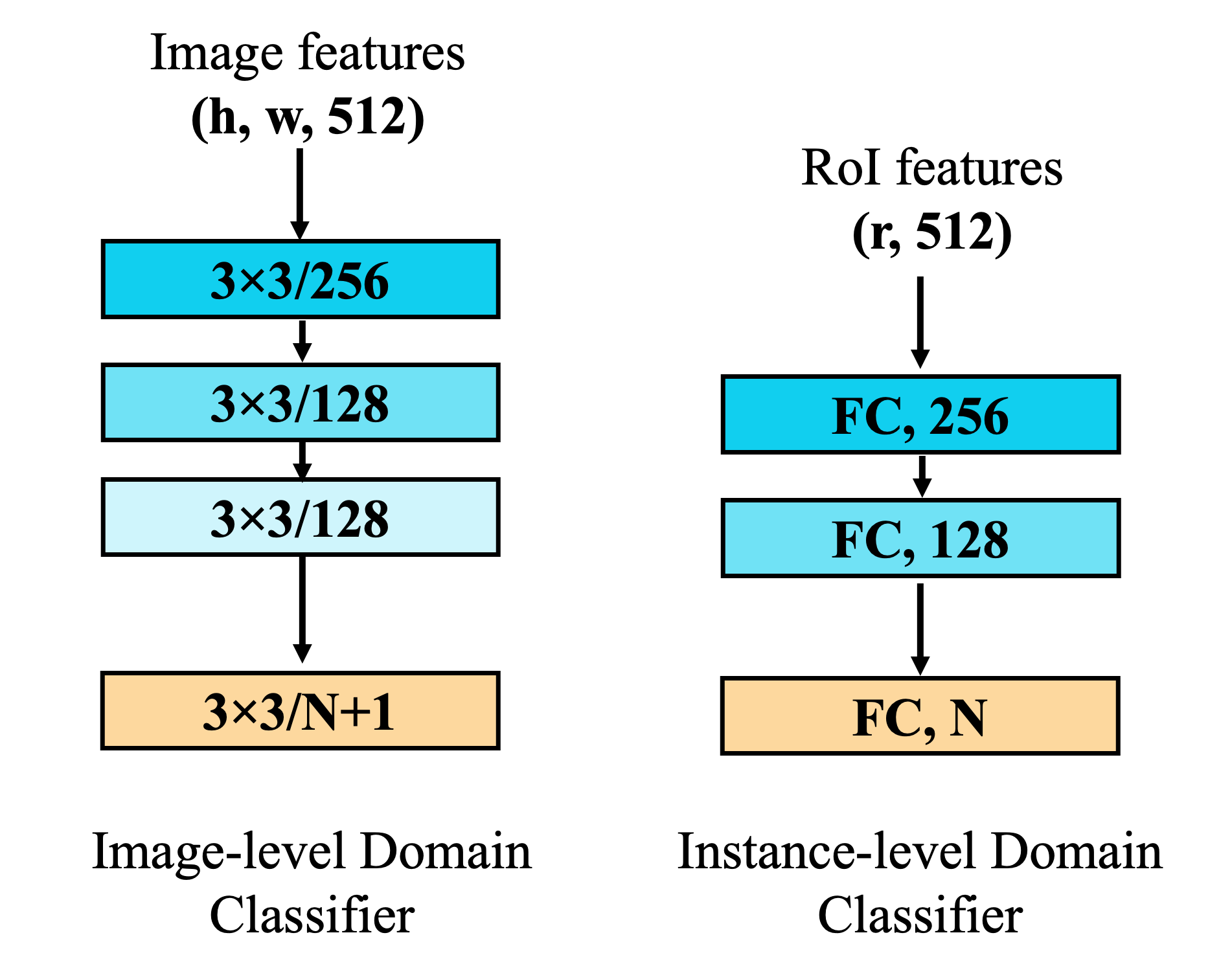}
     {\caption{
      Detailed architecture of the networks used for image-level and instance-level domain classifiers. The activation functions used in image-level and instance-level classifiers are leaky ReLU and GELU respectively. Additionally, Layernorm is used in the instance-level domain classifier. (r= no of GT boxes in the image, h,w = height and width of the feature map, FC = Fully connected layer)} \label{fig:disc_arch}}
\end{figure}

We use two domain classifiers as the discriminator networks for adversarial training: an image-level domain classifier and an instance-level domain classifier. Their architecture is summarized in \cref{fig:disc_arch}. The image-level domain classifier receives its input from the final layer of the backbone network used for feature extraction. This classifier is fully convolutional with a final N+1 class prediction layer (corresponds to the number of source domains plus the target domain). The instance-level classifier receives its input from the attention layer. This classifier consists of only linear layers with a final N-way prediction layer (the target domain is not used here because the GT boxes from the target domain are not available).

\subsection{Class-wise AP} 
We also report the detailed class-wise AP of ACIA on the Cross-Time and Mixed Adaptation settings in \cref{table:Classwise_bdd100k} and \cref{table:Classwise_mixed} respectively. Note that, for the Cross-camera Adaptation settings, there is only one class, so this is not applicable. 

In both settings, our method outperforms the others for all classes. The class-wise AP shows improvement both in the majority and minority classes as well - eg: car (majority) and traffic sign (minority) in the cross-time settings. In the mixed adaptation settings, we outperform others for all the classes, in both the case of two and three source domains. 

\begin{table*}
    \centering
    \resizebox{.99\textwidth}{!}{
    \setlength{\tabcolsep}{3.5pt}
      \begin{tabular}{l | c | c || c c c c c c c c c c | c }
      \hline
    \textbf{Setting} & \textbf{Source} & \textbf{Method}  & \textbf{Bike} & \textbf{Bus} & \textbf{Car} &
    \textbf{Motor} & \textbf{Person} &
    \textbf{Rider} & \textbf{Light} & \textbf{Sign} & \textbf{Train} & \textbf{Truck} & \textbf{mAP}\\ 
    \hline
    \hline
     \multirow{3}{*}{Lower Bound} & D & \multirow{3}{*}{Source Only} & 35.1 & 51.7 & 52.6 & 9.9 & 31.9 & 17.8 & 21.6 & 36.3 & - & 47.1 & 30.4
     \\
     & N &  & 27.9 & 32.5 & 49.4 & 15.0 & 28.7 & 21.8 & 14.0 & 30.5 & - & 30.7 & 25.0
     \\
     & D+N &  & 31.5 & 46.9 & 52.9 & 8.4 & 29.5 & 21.6 & 21.7 & 34.3 & - & 42.2 & 28.9
     \\
     \hline

     \multirow{5}{*}{UDA} & \multirow{5}{*}{D+N} & Strong-Weak\cite{strongweak} & 29.7 & 50.0 & 52.9 & 11.0 & 31.4 & 21.1 & 23.3 & 35.1 & - & 44.9 & 29.9
     \\
     &  & Graph Prototype\cite{MT_graph} & 31.7 & 48.8 & 53.9 & 20.8 & 32.0 & 21.6 & 20.5 & 33.7 & - & 43.1 & 30.6
     \\
     &  & Cat. Regularization\cite{categorical} & 25.3 & 51.3 & 52.1 & 17.0 & 33.4 & 18.9 & 20.7 & 34.8 & - & 47.9 & 30.2
     \\
     &  & UMT\cite{MTunbiased} & 42.3 & 48.1 & 56.4 & 13.5 & 35.3 & 26.9 & 31.1 & 41.7 & - & 40.1 & 33.5
     \\
     &  & Adaptive Teacher\cite{adaptive_teacher} & 43.1 & 48.9 & 56.9 & 14.7 & 36.0 & 27.1 & 32.7 & 43.8 & - & 42.7 & 34.6
     \\
     \hline

     \multirow{6}{*}{MSDA} & \multirow{6}{*}{D+N} & MDAN\cite{MDAN} &  37.1 & 29.9 & 52.8 & 15.8 & 35.1 & 21.6 & 24.7 & 38.8 & - & 20.1 & 27.6
     \\
     & & M$^3$SDA\cite{moment} & 36.9 & 25.9 & 51.9 & 15.1 & 35.7 & 20.5 & 24.7 & 38.1 & - & 15.9 & 26.5
     \\
     & & DMSN\cite{dmsn} & 36.5 & 54.3 & 55.5 & 20.4 & 36.9 & 27.7 & 26.4 & 41.6 & - & 50.8 & 35.0
     \\
     & & TRKP\cite{trkp} & 48.4 & 56.3 & 61.4 & 22.5 & 41.5 & 27.0 & 41.1 & 47.9 & - & 51.9 & 39.8
     \\
     &  & PMT\cite{pmt} & 55.3  & 59.8  & 67.6  &  29.9  &  47.6 & 32.7  &  46.3   & 56.0  &  -  &  57.7 & 45.3
     \\
     \rowcolor{mygray}&  & \textbf{ACIA(Ours)} & \textbf{56.1}  & \textbf{61.0}  & \textbf{69.2}  &  \textbf{31.9}  &  \textbf{51.8} & \textbf{39.8}  &  \textbf{49.2}   & \textbf{59.0}  &  -  &  \textbf{61.0} & \textbf{47.9}
     \\

    \hline
     \multirow{3}{*}{Upper Bound} & \multirow{3}{*}{D+N} & Target Only & 27.2  & 39.6  & 51.9   &  12.7  &  29.0 & 15.2  &  20.0   & 33.1  &  -  &  37.5 & 26.6
     \\
      & &  All-Combined & 56.4  & 59.9  & 67.3  &  30.8  &  47.9 & 33.9  &  47.2   & 57.8  &  -  &  54.8 & 45.3
      \\
      & &  Fine-Tuning & 63.3  & 68.1  & 72.5  &  39.3  &  52.2 & 37.2  &  54.1   & 63.1  &  -  &  59.1 & 50.9
      \\
     \hline
    \end{tabular}
    }
    \caption{Class-wise AP of ACIA compared against the baseline lower bound, UDA, MSDA, and upper bound methods in the cross-time settings. Source domains are daytime (D) and night (N) subsets and the target is always Dusk/Dawn of BDD100K.}
    \label{table:Classwise_bdd100k}
\end{table*}

\begin{table*}
    \centering
    \resizebox{.90\textwidth}{!}{
    \setlength{\tabcolsep}{3.5pt}
      \begin{tabular}{l | c | c || c c c c c c c  | c }
      \hline
    \textbf{Setting} & \textbf{Source} & \textbf{Method}  & \textbf{Person} & \textbf{Car}  &
    \textbf{Rider} & \textbf{Truck} &
    \textbf{Motor} & \textbf{Bicycle} & \textbf{Bike} & \textbf{mAP}\\ 
    \hline
    \hline
     Lower Bound & C & Source Only  & 26.9 & 44.7 & 22.1 & 17.4 & 17.1 & 18.8 & 16.7 & 23.4
     \\
    \hline
     Lower Bound &\multirow{8}{*} {C+M} & Source Only & 35.2 & 49.5 & 26.1 & 25.8 & 18.9 & 26.1 & 26.5 & 29.7
     \\
    UDA  &  & UMT\cite{MTunbiased} & 30.7 & 28.0 & 3.9 & 11.2 & 19.2 & 17.8 & 18.7 & 18.5
    \\
    UDA &  & Adaptive Teacher\cite{adaptive_teacher} & 31.2 & 31.7 & 15.1 & 16.4 & 17.1 & 20.9 & 27.9 & 22.9
    \\
    MSDA &  &  TRKP\cite{trkp} & 39.2 & 53.2 & 32.4 & 28.7 & 25.5 & 31.1 & 37.4 & 35.3
     \\
     MSDA &  & PMT\cite{pmt} & 41.1 & 53.5 & 31.2 & 31.9 & 33.7 & 34.9 & 44.6 & 38.7
     \\
     \rowcolor{mygray} MSDA &  & \textbf{ACIA(ours)} & \textbf{43.3} & \textbf{58.1} & \textbf{33.3} & \textbf{35.1} & \textbf{33.7} & \textbf{38.6} & \textbf{45.2} & \textbf{41.0}
     \\
     Upper Bound &  & All-Combined & 40.2 & 60.1 & 47.1 & 60.0 & 29.2 & 36.3 & 56.9 & 47.1
     \\
     Upper Bound &  & Fine-Tuning & 44.1 & 61.4 & 49.0 & 61.1 & 30.8 & 39.2 & 58.8 & 49.2
     \\
    \hline
     
      Lower Bound & \multirow{8}{*}{C+M+S} & Source Only & 36.6 & 49.0 & 22.8 & 24.9 & 26.9 & 28.4 & 27.7 & 30.9
     \\
      UDA &  & UMT\cite{MTunbiased} & 32.7 & 39.6 & 6.6 & 21.2 & 21.3 & 25.7 & 28.5 & 25.1
     \\
     UDA &  & Adaptive Teacher\cite{adaptive_teacher} & 36.3 & 42.6 & 19.7 & 23.4 & 24.8 & 27.1 & 33.2 & 29.6
     \\
      MSDA &  &  TRKP\cite{trkp} & 40.2 & 53.9 & 31.0 & 30.8 & 30.4 & 34.0 & 39.3 & 37.1
     \\
     MSDA &  & PMT\cite{pmt} & 43.3 & 54.1 & 32.0 & 32.6 & 35.1 & 36.1 & 44.8 & 39.7
     \\
     \rowcolor{mygray} MSDA &  & \textbf{ACIA(ours)} & \textbf{44.9} & \textbf{59.2} & \textbf{33.8} & \textbf{33.5} & \textbf{38.3} & \textbf{39.9} & \textbf{46.5} & \textbf{42.3}
     \\
      Upper Bound &  & All-Combined & 41.7 & 63.9 & 49.5 & 58.1 & 31.6 & 39.1 & 53.5 & 48.2
      \\
      Upper Bound &  & Fine-Tuning & 49.2 & 63.5 & 56.1 & 62.6 & 35.1 & 43.7 & 57.2 & 52.5
    
     \\
     \hline

     Upper Bound & C+M+S &  Target Only & 35.3 & 53.9 & 33.2 & 46.3 & 25.6 & 29.3 & 46.7 & 38.6

     \\
     \hline
    \end{tabular}
    }
    \caption{Class-wise AP of ACIA compared against the baselines in the mixed adaptation settings. Source domains are Cityscapes(C), MS COCO(M), and Synscapes(S) datasets while the Daytime domain of BDD100K is the target domain.}
    \label{table:Classwise_mixed}
\end{table*}

\subsection{More Detection Visualization}
\begin{figure*}[!ht]
\centering
\begin{tabular}{c @{\hspace{-0.3cm}} c  @{\hspace{-0.3cm}} c}
\includegraphics[width=0.33\linewidth]{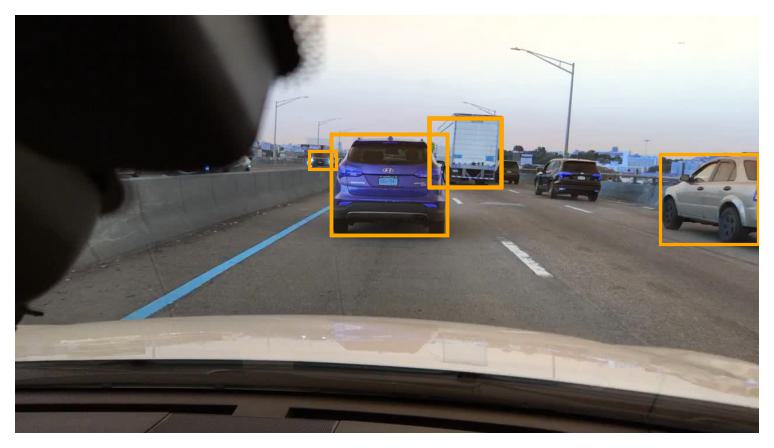} &
\includegraphics[width=0.33\linewidth]{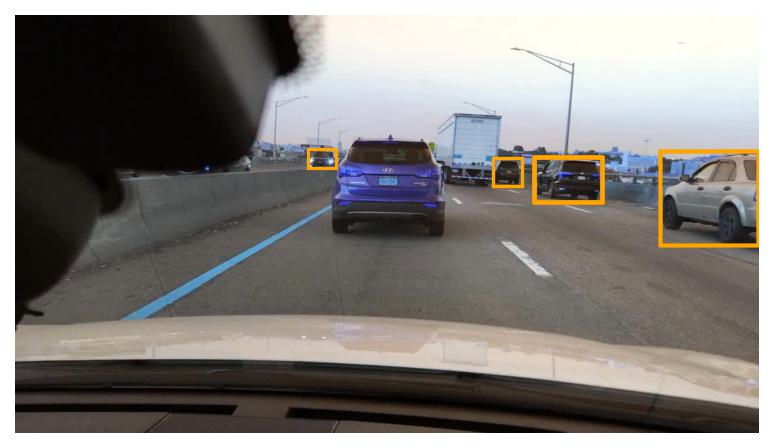} &
\includegraphics[width=0.33\linewidth]{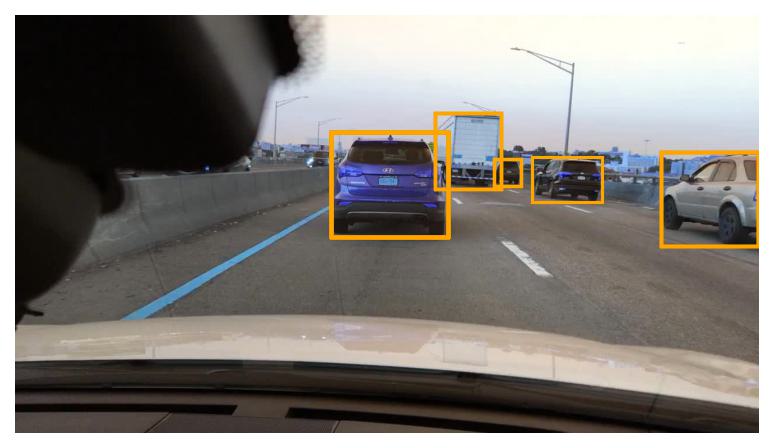} \\
\includegraphics[width=0.33\linewidth]{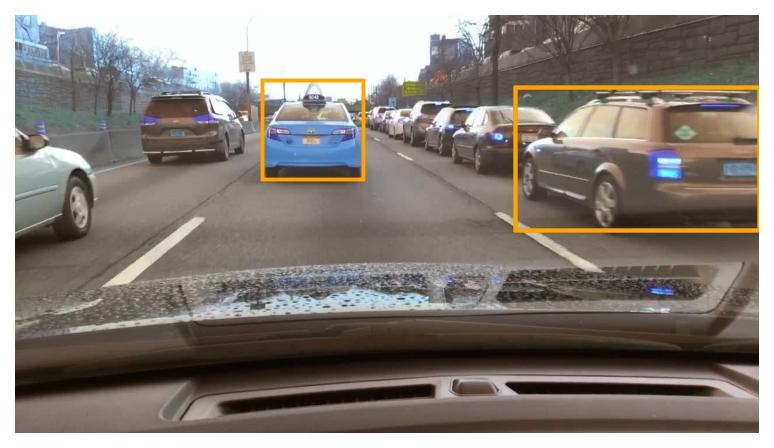} &
\includegraphics[width=0.33\linewidth]{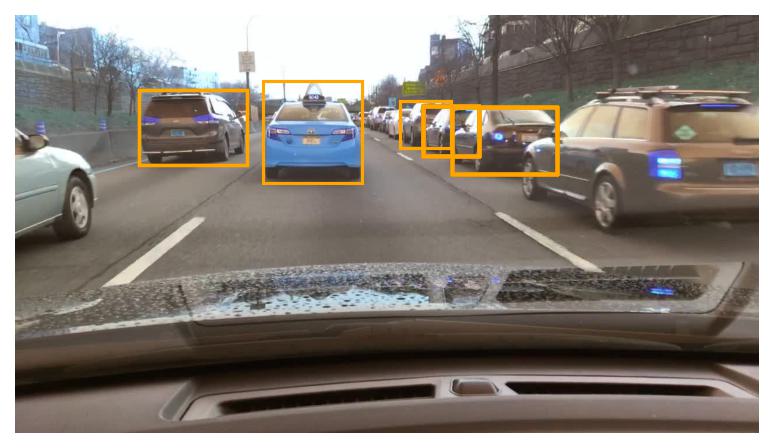} &
\includegraphics[width=0.33\linewidth]{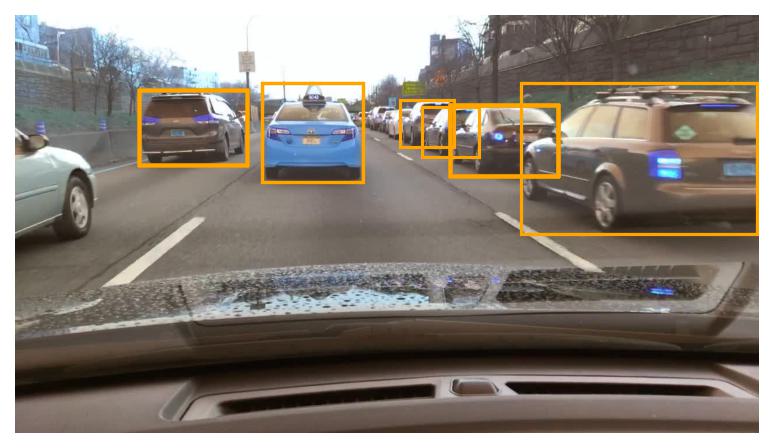} \\
\includegraphics[width=0.33\linewidth ]{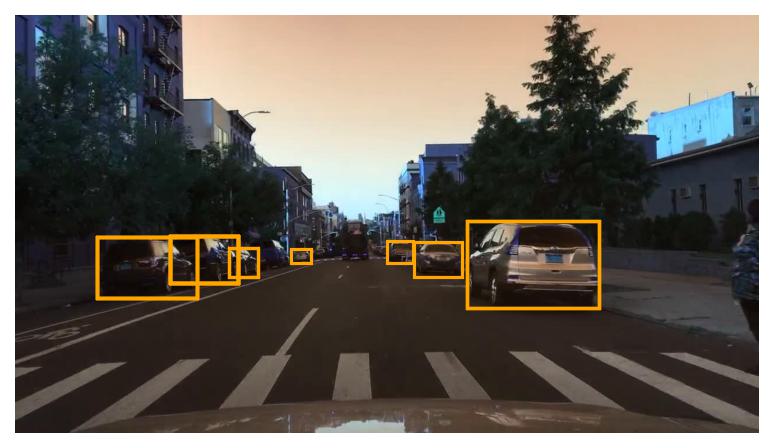} &
\includegraphics[width=0.33\linewidth ]{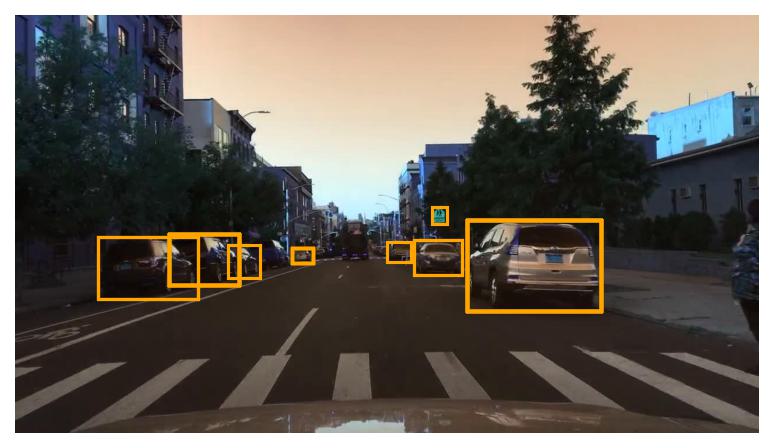} &
\includegraphics[width=0.33\linewidth ]{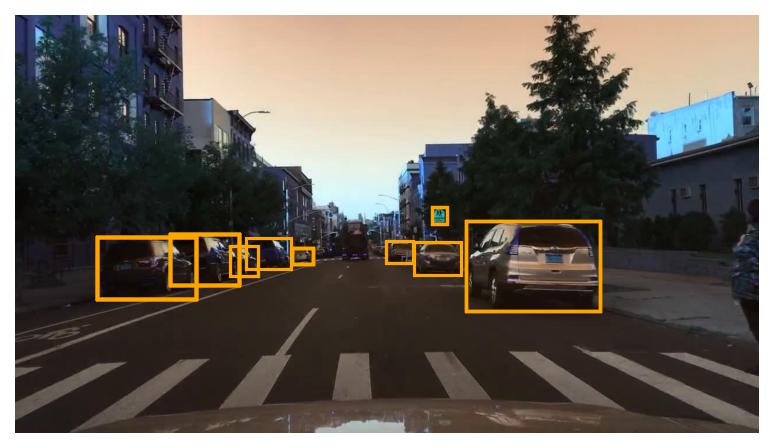} \\
\includegraphics[width=0.33\linewidth]{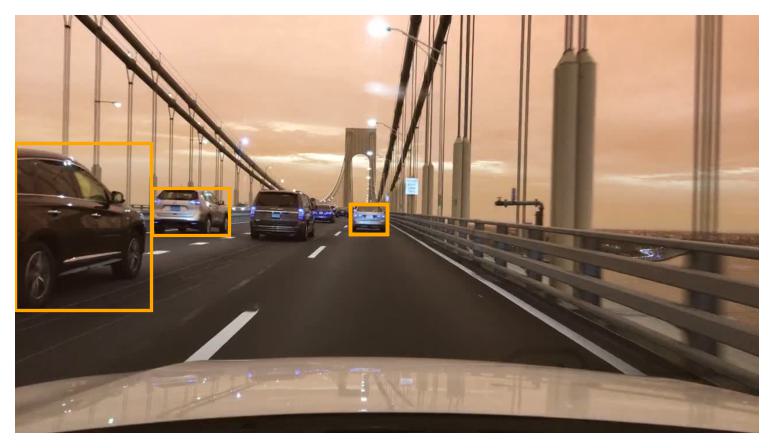} &
\includegraphics[width=0.33\linewidth ]{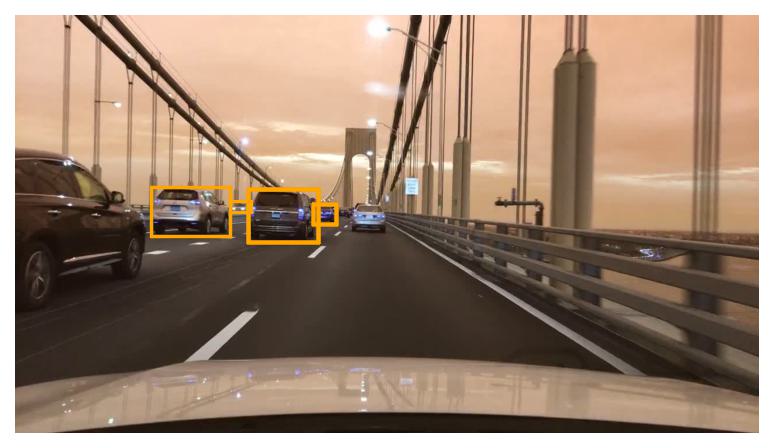} &
\includegraphics[width=0.33\linewidth ]{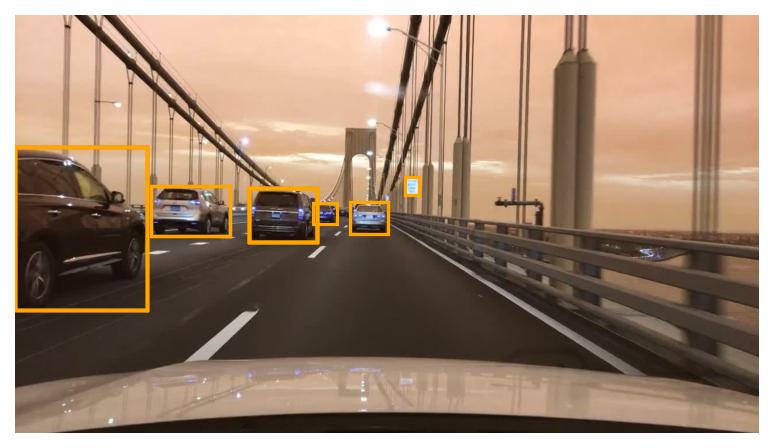} \\
\includegraphics[width=0.33\linewidth ]{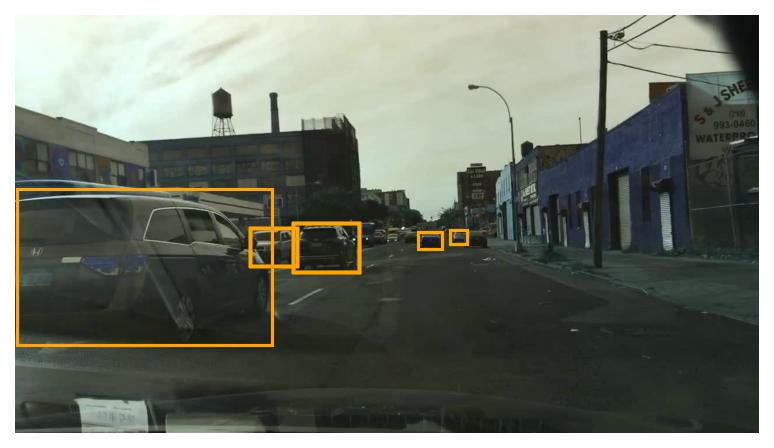} &
\includegraphics[width=0.33\linewidth ]{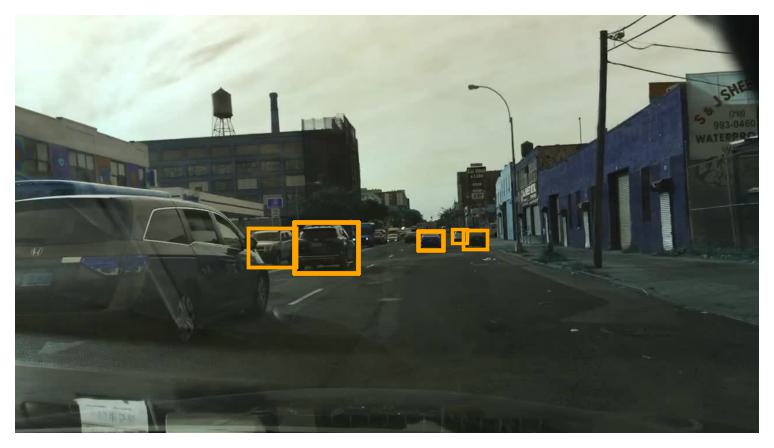} &
\includegraphics[width=0.33\linewidth ]{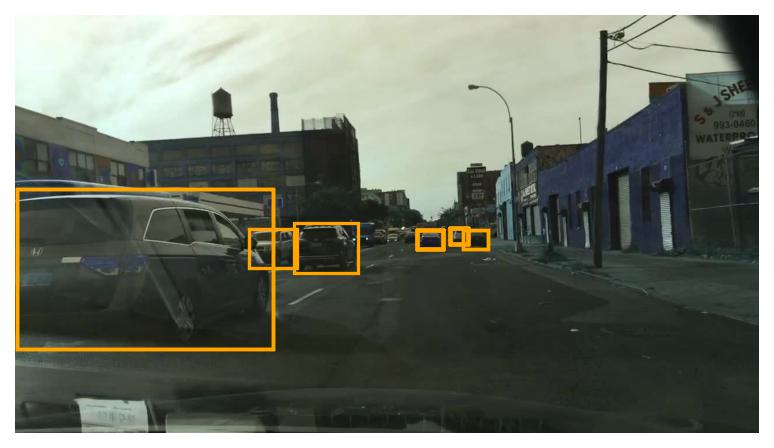} \\
\includegraphics[width=0.33\linewidth ]{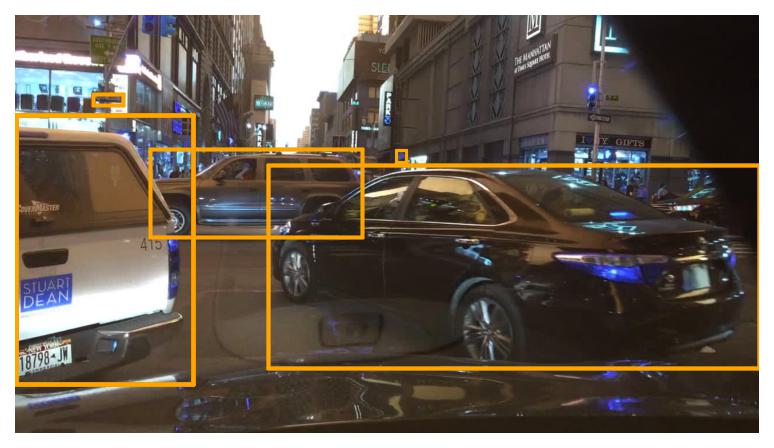} &
\includegraphics[width=0.33\linewidth ]{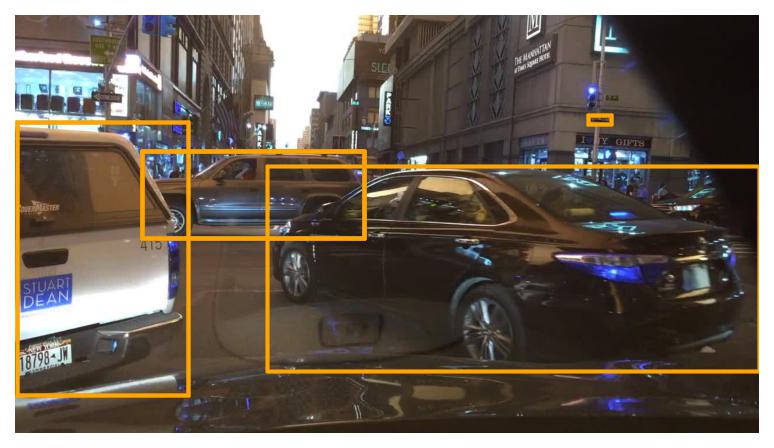} &
\includegraphics[width=0.33\linewidth ]{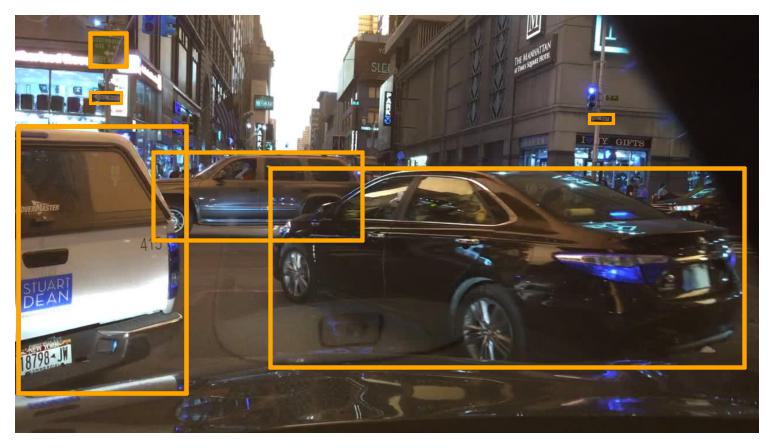} \\
 (a) Methods w/o class alignment.  & (b) Prototype-based method. & (c) Our ACIA method. \\
\end{tabular}
 \caption{Comparison of instance-level adaptation detection on BDD100k cross-time setting. \textbf{(a)} MSDA without class-conditioned instance adaptation as in \cite{dmsn, trkp}. \textbf{(b)} With the prototype-based class-conditional adaptation \cite{pmt}. \textbf{(c)} ODs with our ACIA approach.}
 \label{fig:sup_detections}
\end{figure*}

In \cref{fig:sup_detections} we present more visualization of the detections on BDD100k cross-time for the three multi-source adaptation approaches presented in the paper: no class-wise adaptation (similar to \cite{dmsn,trkp}), prototype-based class-conditional adaptation\cite{pmt} and our attention-based class-conditional adaptation. From the visualizations, it can be observed that our method is performing better detection compared to the other approaches highlighting the impact of an efficient class-conditioned alignment. 

\clearpage


%
%
\clearpage
\bibliographystyle{splncs04}
\bibliography{egbib}


%
%
\clearpage
\bibliographystyle{ieee_fullname}
\bibliography{egbib}

\begin{thebibliography}{10}
\providecommand{\url}[1]{\texttt{#1}}
\providecommand{\urlprefix}{URL }
\providecommand{\doi}[1]{https://doi.org/#1}

\bibitem{coupled-training-msda}
Amosy, O., Chechik, G.: Coupled training for multi-source domain adaptation. In: IEEE/CVF Winter Conference on Applications of Computer Vision (WACV) (2022)

\bibitem{atten1}
Bahdanau, D., Cho, K., Bengio, Y.: Neural machine translation by jointly learning to align and translate (2016)

\bibitem{pmt}
Belal, A., Meethal, A., Romero, F.P., Pedersoli, M., Granger, E.: Multi-source domain adaptation for object detection with prototype-based mean teacher. In: IEEE/CVF Winter Conference on Applications of Computer Vision (WACV). pp. 1277--1286 (January 2024)

\bibitem{MT_graph}
Cai, Q., Pan, Y., Ngo, C.W., Tian, X., Duan, L., Yao, T.: Exploring object relation in mean teacher for cross-domain detection (2019). \doi{10.48550/ARXIV.1904.11245}, \url{https://arxiv.org/abs/1904.11245}

\bibitem{detr}
Carion, N., Massa, F., Synnaeve, G., Usunier, N., Kirillov, A., Zagoruyko, S.: End-to-end object detection with transformers (2020)

\bibitem{error_accumulation}
Chen, C., Xie, W., Huang, W., Rong, Y., Ding, X., Huang, Y., Xu, T., Huang, J.: Progressive feature alignment for unsupervised domain adaptation (2019)

\bibitem{HTCN}
Chen, C., Zheng, Z., Ding, X., Huang, Y., Dou, Q.: Harmonizing transferability and discriminability for adapting object detectors (2020)

\bibitem{wildDA}
Chen, Y., Li, W., Sakaridis, C., Dai, D., Van~Gool, L.: Domain adaptive faster r-cnn for object detection in the wild (2018). \doi{10.48550/ARXIV.1803.03243}, \url{https://arxiv.org/abs/1803.03243}

\bibitem{cityscapes}
Cordts, M., Omran, M., Ramos, S., Rehfeld, T., Enzweiler, M., Benenson, R., Franke, U., Roth, S., Schiele, B.: The cityscapes dataset for semantic urban scene understanding (2016)

\bibitem{MTunbiased}
Deng, J., Li, W., Chen, Y., Duan, L.: Unbiased mean teacher for cross-domain object detection (2021)

\bibitem{GRL}
Ganin, Y., Lempitsky, V.: Unsupervised domain adaptation by backpropagation (2014)

\bibitem{kitty}
Geiger, A., Lenz, P., Urtasun, R.: Are we ready for autonomous driving? the kitti vision benchmark suite. In: 2012 IEEE Conference on Computer Vision and Pattern Recognition. pp. 3354--3361 (2012). \doi{10.1109/CVPR.2012.6248074}

\bibitem{maskrcnn}
He, K., Gkioxari, G., Dollár, P., Girshick, R.: Mask r-cnn (2018)

\bibitem{MCAR}
He, Z., Zhang, L.: Multi-adversarial faster-rcnn for unrestricted object detection (2019)

\bibitem{progressive}
Hsu, H.K., Yao, C.H., Tsai, Y.H., Hung, W.C., Tseng, H.Y., Singh, M., Yang, M.H.: Progressive domain adaptation for object detection (2019). \doi{10.48550/ARXIV.1910.11319}, \url{https://arxiv.org/abs/1910.11319}

\bibitem{tradeoff-da}
Kundu, J.N., Kulkarni, A.R., Bhambri, S., Mehta, D., Kulkarni, S.A., Jampani, V., Radhakrishnan, V.B.: Balancing discriminability and transferability for source-free domain adaptation. In: International Conference on Machine Learning (2022)

\bibitem{adaptive_teacher}
Li, Y.J., Dai, X., Ma, C.Y., Liu, Y.C., Chen, K., Wu, B., He, Z., Kitani, K., Vajda, P.: Cross-domain adaptive teacher for object detection (2021). \doi{10.48550/ARXIV.2111.13216}, \url{https://arxiv.org/abs/2111.13216}

\bibitem{retinanet-focal-loss}
Lin, T.Y., Goyal, P., Girshick, R., He, K., Dollár, P.: Focal loss for dense object detection. IEEE Transactions on Pattern Analysis and Machine Intelligence  (2020)

\bibitem{mscoco}
Lin, T.Y., Maire, M., Belongie, S., Bourdev, L., Girshick, R., Hays, J., Perona, P., Ramanan, D., Zitnick, C.L., Dollár, P.: Microsoft coco: Common objects in context (2015)

\bibitem{groundingdino-liu-2023}
Liu, S., Zeng, Z., Ren, T., Li, F., Zhang, H., Yang, J., Li, C., Yang, J., Su, H., Zhu, J., et~al.: Grounding dino: Marrying dino with grounded pre-training for open-set object detection. arXiv preprint arXiv:2303.05499  (2023)

\bibitem{atten2}
Luong, T., Pham, H., Manning, C.D.: Effective approaches to attention-based neural machine translation. In: M{\`a}rquez, L., Callison-Burch, C., Su, J. (eds.) Proceedings of the 2015 Conference on Empirical Methods in Natural Language Processing. pp. 1412--1421. Association for Computational Linguistics, Lisbon, Portugal (Sep 2015). \doi{10.18653/v1/D15-1166}, \url{https://aclanthology.org/D15-1166}

\bibitem{moment}
Peng, X., Bai, Q., Xia, X., Huang, Z., Saenko, K., Wang, B.: Moment matching for multi-source domain adaptation (2019)

\bibitem{fasterrcnn}
Ren, S., He, K., Girshick, R., Sun, J.: Faster r-cnn: Towards real-time object detection with region proposal networks (2016)

\bibitem{styletransfer}
Rodriguez, A.L., a, Mikolajczyk, K.: Domain adaptation for object detection via style consistency (2019). \doi{10.48550/ARXIV.1911.10033}, \url{https://arxiv.org/abs/1911.10033}

\bibitem{Weak-strong}
Saito, K., Ushiku, Y., Harada, T., Saenko, K.: Strong-weak distribution alignment for adaptive object detection (2018). \doi{10.48550/ARXIV.1812.04798}, \url{https://arxiv.org/abs/1812.04798}

\bibitem{strong_weak_alignment}
Saito, K., Ushiku, Y., Harada, T., Saenko, K.: Strong-weak distribution alignment for adaptive object detection. In: CVPR (2019)

\bibitem{strongweak}
Saito, K., Ushiku, Y., Harada, T., Saenko, K.: Strong-weak distribution alignment for adaptive object detection. In: IEEE/CVF Conference on Computer Vision and Pattern Recognition (CVPR) (2019)

\bibitem{eqlv2}
Tan, J., Lu, X., Zhang, G., Yin, C., Li, Q.: Equalization loss v2: A new gradient balance approach for long-tailed object detection (2021), \url{https://arxiv.org/abs/2012.08548}

\bibitem{attention_is_all}
Vaswani, A., Shazeer, N., Parmar, N., Uszkoreit, J., Jones, L., Gomez, A.N., Kaiser, L., Polosukhin, I.: Attention is all you need (2023)

\bibitem{msda-classif-neurips-2020}
Venkat, N., Kundu, J.N., Singh, D.K., Revanur, A., Babu, R.V.: Your classifier can secretly suffice multi-source domain adaptation. In: Advances in Neural Information Processing Systems (NeurIPS) (2020)

\bibitem{megacda}
VS, V., Gupta, V., Oza, P., Sindagi, V.A., Patel, V.M.: Mega-cda: Memory guided attention for category-aware unsupervised domain adaptive object detection (2021)

\bibitem{yolov7-wang-2023}
Wang, C.Y., Bochkovskiy, A., Liao, H.Y.M.: {YOLOv7}: Trainable bag-of-freebies sets new state-of-the-art for real-time object detectors. In: IEEE/CVF Conference on Computer Vision and Pattern Recognition (CVPR) (2023)

\bibitem{synscapes}
Wrenninge, M., Unger, J.: Synscapes: A photorealistic synthetic dataset for street scene parsing (2018)

\bibitem{trkp}
Wu, J., Chen, J., He, M., Wang, Y., Li, B., Ma, B., Gan, W., Wu, W., Wang, Y., Huang, D.: Target-relevant knowledge preservation for multi-source domain adaptive object detection (2022). \doi{10.48550/ARXIV.2204.07964}, \url{https://arxiv.org/abs/2204.07964}

\bibitem{categorical}
Xu, C.D., Zhao, X.R., Jin, X., Wei, X.S.: Exploring categorical regularization for domain adaptive object detection (2020)

\bibitem{graph_induced}
Xu, M., Wang, H., Ni, B., Tian, Q., Zhang, W.: Cross-domain detection via graph-induced prototype alignment (2020)

\bibitem{dmsn}
Yao, X., Zhao, S., Xu, P., Yang, J.: Multi-source domain adaptation for object detection (2021). \doi{10.48550/ARXIV.2106.15793}, \url{https://arxiv.org/abs/2106.15793}

\bibitem{bdd100k}
Yu, F., Chen, H., Wang, X., Xian, W., Chen, Y., Liu, F., Madhavan, V., Darrell, T.: Bdd100k: A diverse driving dataset for heterogeneous multitask learning (2020)

\bibitem{MDAN}
Zhao, H., Zhang, S., Wu, G., Moura, J.M.F., Costeira, J.P., Gordon, G.J.: Adversarial multiple source domain adaptation. In: Advances in Neural Information Processing Systems (2018)

\bibitem{adversarial_msda-neurips-2018}
Zhao, H., Zhang, S., Wu, G., Moura, J.M.F., Costeira, J.P., Gordon, G.J.: Adversarial multiple source domain adaptation. In: Bengio, S., Wallach, H., Larochelle, H., Grauman, K., Cesa-Bianchi, N., Garnett, R. (eds.) Advances in Neural Information Processing Systems. Curran Associates, Inc. (2018)

\bibitem{coursetofine}
Zheng, Y., Huang, D., Liu, S., Wang, Y.: Cross-domain object detection through coarse-to-fine feature adaptation (2020)

\bibitem{detic-zhou-2022}
Zhou, X., Girdhar, R., Joulin, A., Kr{\"a}henb{\"u}hl, P., Misra, I.: Detecting twenty-thousand classes using image-level supervision. In: ECCV (2022)

\bibitem{selective}
Zhu, X., Pang, J., Yang, C., Shi, J., Lin, D.: Adapting object detectors via selective cross-domain alignment. In: IEEE/CVF Conference on Computer Vision and Pattern Recognition (CVPR) (June 2019)

\end{thebibliography}
\end{document}